\documentclass[10pt,twocolumn,letterpaper]{article}

\usepackage{xcolor}

\usepackage{iccv}
\usepackage{times}
\usepackage{epsfig}
\usepackage{graphicx}
\usepackage{amsmath}
\usepackage{amssymb}
\usepackage{amsfonts}
\usepackage{algorithmic}
\usepackage{algorithm}
\usepackage{array}
\usepackage[caption=false]{subfig}
\usepackage{textcomp}
\usepackage{stfloats}
\usepackage{url}
\usepackage{verbatim}
\usepackage{graphicx}
\usepackage{cite}
\usepackage{bm}
\usepackage{breqn,xspace}
\usepackage{float}
\usepackage{ragged2e}
\usepackage{color}
\usepackage{xparse}
\usepackage{amsopn}
\usepackage[outdir=./pic/]{epstopdf}

\usepackage[pagebackref=true,breaklinks=true,letterpaper=true,colorlinks,bookmarks=false]{hyperref}

\iccvfinalcopy 

\def\iccvPaperID{9910} 

\def\iccvPaperID{9910} %

\ificcvfinal\fi

\ifodd 0
\newcommand{\rev}[1]{{\color{blue}#1}} %
\newcommand{\newrev}[1]{{\color{red}#1}} %
\else
\newcommand{\rev}[1]{#1}
\newcommand{\newrev}[1]{#1} %
\fi

\ifodd 0
\newcommand{\camrev}[1]{{\color{cyan}#1}} %
\else
\newcommand{\camrev}[1]{#1}

\newcommand{\name}{OCHID-Fi\xspace}
\newcommand{\nname}{OCH-Net\xspace}
\newcommand{\nnname}{OCH-AL\xspace}

\ificcvfinal\fi

\begin{document}

\title{\name: Occlusion-Robust Hand Pose Estimation in 3D via RF-Vision}

\author{Shujie Zhang$^{1}$\thanks{Equal contribution.  https://github.com/DeepWiSe888/OCHID-Fi},~~Tianyue Zheng$^{1}$\footnotemark[1],~~Zhe Chen$^{2}$,~~Jingzhi Hu$^{1}$, \\
Abdelwahed Khamis$^{3}$, Jiajun Liu$^{3}$,~~Jun Luo$^{1}$ \\
$^1$ Nanyang Technological University \quad
$^2$ Fudan University \quad
$^3$ CSIRO\\
{\tt\small \{shujie002,tianyue002,jingzhi.hu,junluo\}@ntu.edu.sg}
\\
{\tt\small zhechen@fudan.edu.cn\quad}  
{\tt\small abdelwahed.khamis@data61.csiro.au\quad}  
{\tt\small jiajun.liu@csiro.au}
}

\maketitle
\ificcvfinal\thispagestyle{empty}\fi

\begin{abstract}
Hand Pose Estimation (HPE) is crucial to many applications, 
but conventional \underline{c}a\underline{m}eras-based CM-HPE methods are completely subject to Line-of-Sight (LoS), as cameras cannot capture occluded objects. In this paper, we propose to exploit Radio-Frequency-Vision (RF-vision) capable of bypassing obstacles for achieving occluded HPE, and we introduce \name as the first RF-HPE method with 3D pose estimation capability.  
\name employs wideband RF sensors widely available on smart devices (e.g., iPhones) to probe 3D human hand pose and extract their skeletons behind obstacles. To overcome the challenge in labeling RF imaging given its human incomprehensible nature, \name employs a cross-modality and cross-domain training process.
It uses a pre-trained CM-HPE network and a synchronized CM/RF dataset, to guide the training of its complex-valued RF-HPE network under LoS conditions. 
It further transfers knowledge learned from labeled LoS domain to unlabeled occluded domain via adversarial learning, enabling \name to generalize to unseen occluded scenarios.
Experimental results demonstrate the superiority of \name:
it achieves comparable accuracy to CM-HPE under normal conditions while maintaining such accuracy even in occluded scenarios, with empirical evidence for its generalizability to new domains.
\end{abstract}
\vspace{-2ex}
\section{Introduction}

We have witnessed tremendous efforts put into Computer Vision (CV) research in the past decade, driven by applications such as facial recognition~\cite{wang2018orthogonal,Song_2019_ICCV}, hand pose estimation~\cite{moon2020interhand2, spurr2021self}, object detection~\cite{zhu2017couplenet, jiang2018acquisition}, and augmented/virtual reality~\cite{zhao2020pointar,kim2019deep}. Among various sensing technologies behind CV, Optical Vision (OV) has so far been the dominant path, fuelled by the widely available OV devices (i.e., cameras, lidars) and large-scale  datasets~\cite{deng2009imagenet, kosaka2000augmented}. However, OV often suffers from a few major limiting factors: it requires Line-of-Sight (LoS)~\cite{deng2009imagenet, kosaka2000augmented} and certain lighting conditions~\cite{arici2009histogram,Xu2020low_light}, it is prone to background clutter~\cite{konstantinou2019adaptive,sun2006background}, and is challenged by privacy concerns~\cite{ascl, sclrr}.

Motivated by these shortcomings, non-optical vision technologies have also been explored in the CV community~\cite{Ouaknine_2021_ICCV,oralkan2002capacitive,wang2019person, ye2021channel,zhao2018through,qian2021robust,radarnet2020,zheng2021siwa,ding2020rf,chen2021rf,zheng2022SoM}, among which \textit{Radio-Frequency}-vision (\textit{RF-vision}) stands out in many aspects including low complexity (thus real-time responsiveness), energy efficiency, and ready deployability~\cite{chen2021octopus, Ouaknine_2021_ICCV}.
These strengths have motivated the use of RF-vision for problems previously tackled by OV~\cite{qian2021robust,wang2019person,zhao2018through}. 
However, one of the biggest strengths of RF-vision, i.e., \textit{occlusion robustness}, has only been lightly touched by RF-Pose~\cite{zhao2018through}, for coarse-grained human pose estimation trained only in LoS domain and used directly in obstructed scenarios without accounting for the impact of the obstacles that create occlusion. 

\begin{figure}[t]
    \vspace{-1.5ex}
    \setlength\abovecaptionskip{8pt}
	\centering
	\includegraphics[width=0.45\textwidth]{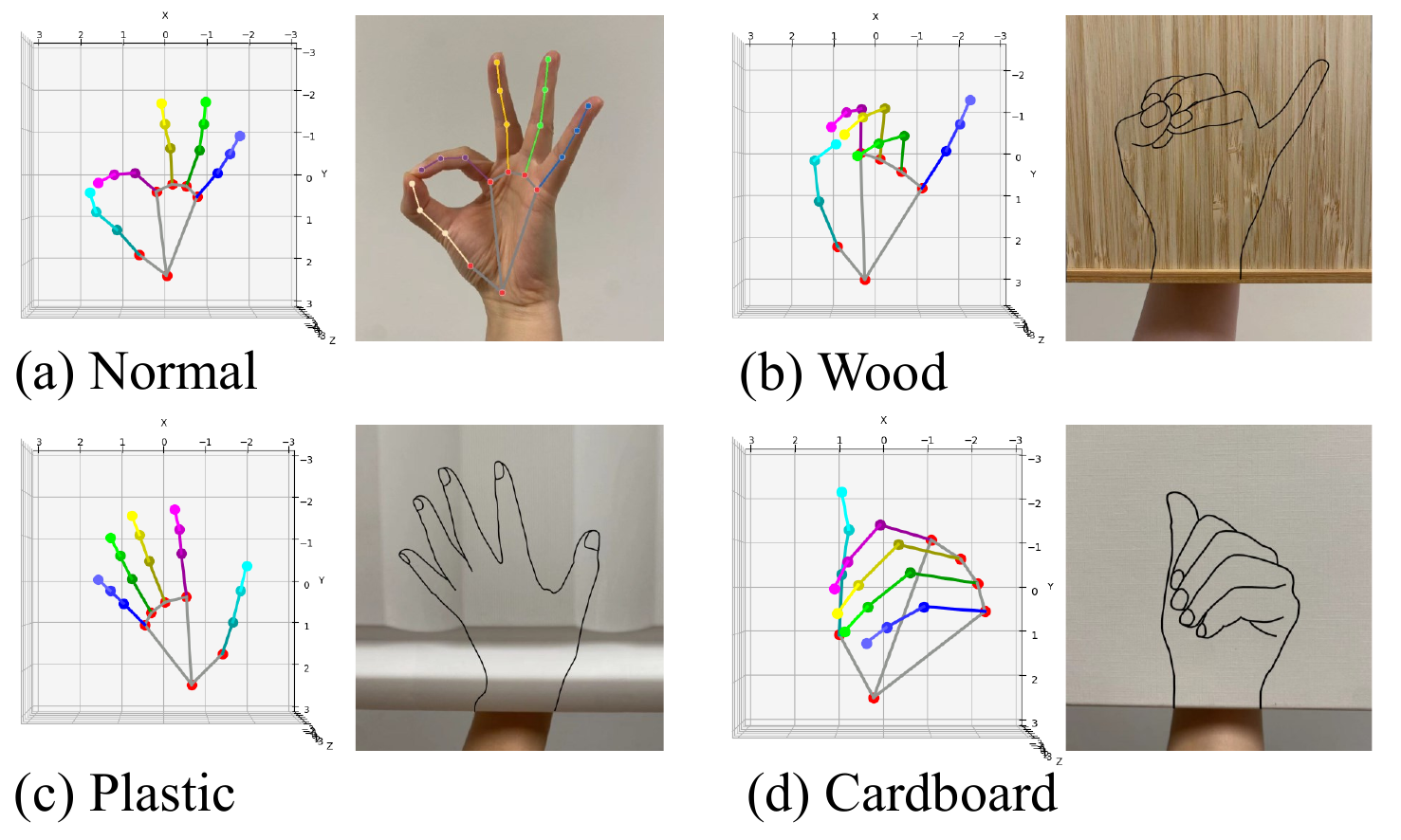}
	\caption{Unlike CM-HPE models, \name can extract 3D hand keypoints behind occlusions (b-d). Note that the wrist and the position of the hand are shown here only as an example. In the actual experiment, the hand can be fully occluded and its position can vary. Drawn outlines here are for illustration purposes only.}
	\label{fig:teaser}
     \vspace{-2ex}
\end{figure}

In this paper, we focus on utilizing RF-vision to perform fine-grained 3D \textit{Hand Pose Estimation} (HPE) for occluded scenes.  
It is important to distinguish RF-HPE from RF-enabled \textit{Hand Gesture Recognition} (RF-HGR)~\cite{zheng2019zero,venkatnarayan2018multi}: While the former requires a more detailed understanding of hand keypoints~\cite{moon2020interhand2,spurr2018cross,zhang2020mediapipe}, the latter only performs basic classification tasks. 
As a result, RF-HPE is highly non-trivial and we summarize three major challenges:

\begin{itemize}
    \vspace{-1.5ex}
    \item \textsl{Non-Euclidean Mapping of Keypoints:} 
    RF data does not enjoy a direct Euclidean mapping from signal space to keypoint locations. It is hence difficult for a deep learning model to uncover the intrinsic relations.
    \item \textsl{Model Design for Low-resolution, Complex-valued RF data:}  
    CM-HPE models cannot process low-resolution complex-valued RF data, while RF-HGR models are designed exclusively for classification while RF-HPE is inherently a regression task. 
    \vspace{-1ex}
    \item \textsl{Model Training for Occluded Scenes:} RF data in occluded scenes are significantly affected by reflection and refraction caused by the obstacles. It is highly non-trivial to design a training mechanism for an RF-HPE model in such scenarios.
\vspace{-1ex}
\end{itemize}

To address these challenges, we propose \name, the first 3D RF-HPE model capable of extracting 3D hand keypoints behind full occlusion. To provide a taste of what \name can achieve, we plot the outputs of \name and a state-of-the-art CM-HPE solution (Google MediaPipe Hands~\cite{zhang2020mediapipe}) for a clear comparison in Figure~\ref{fig:teaser}.
Trained in \textit{cross-modality} and \textit{cross-domain} manner, \name exploits RF-vision to tackle the occlusion issue where CM-HPE fails. 
\name translates RF signals to hand keypoints through cross-modality training aided by a pre-trained CM-HPE model. 
Specifically, \name employs a synchronized pair of camera and RF sensor during data collection in LoS scenarios.
The CM-HPE network is first trained with pseudo ground truth collected by the OptiTrack~\cite{nagymate2018application} system, and then we transfer its learned knowledge to \nname by supervising \nname together with the RF ground truth data. 
To handle the complex-valued RF data, a deep complex-valued network is specifically designed to perform feature extraction.
While completing the first training stage allows 
\camrev{the deep complex-valued network} \nname~\camrev{(OCcluded Hand-Net)} to work independently under LoS situation, 
the second stage training \nnname~\camrev{(OCcluded Hand-Adversarial Learning)} is performed to 
further transfer knowledge across domains (from LoS to occluded), for which we leverage adversarial learning in an unsupervised manner.
\vspace{1ex}
In summary, our major contributions are: 

\begin{itemize}
    \vspace{-1ex}
    \item To the best of our knowledge, \name is the first occlusion-robust 3D RF-HPE model.
    \vspace{-1.2ex}
    \item \name transfers the knowledge from the OV to the RF, effectively bridging the gap between complex RF-vision data and hand keypoints.
    \vspace{-1.2ex}
    \item \nname is proposed as the complex-valued RF-HPE model to fit RF signals, making it possible to fully exploit the intrinsic RF features. 
    \vspace{-1.2ex}
    \item \nnname leverages adversarial learning in an unsupervised manner, so as to further transfer knowledge from the LoS domain into the occluded one.
    \vspace{-1.2ex}
    \item We perform extensive experiments to validate that, in occluded scenes where OV fails completely, \name achieves similar accuracy to that of CM-HPE in LoS conditions. Our empirical results also demonstrate that \name generalizes to unseen occluded scenes.
\end{itemize}

\section{Related Work}  \label{sec:rw} %
\label{sec:rel}
Several OV methods exist for HGR~\cite{zhang2017Gesture, Cao_2017_ICCV, nguyen2019neural, min2020efficient} using photos and videos. However, their functions cannot meet the need from the HPE problem, as HGR aims to only classify hand gestures rather than estimate locations of hand keypoints (such as knuckles) accurately. To this end, novel solutions for addressing the HPE problem have been devised based on visual inputs~\cite{ge20193d,spurr2021self,cheng2021handfold,zhang2020mediapipe,iqbal2018hand,simon2017hand}. Specifically, HandFoldingNet~\cite{cheng2021handfold} uses  depth image as input, OpenPose~\cite{simon2017hand} employs a multi-camera system for fine-grained hand keypoint detection, and   others~\cite{ge20193d,spurr2021self,zhang2020mediapipe,iqbal2018hand} rely on neural networks for extracting hand keypoint out of single RGB images. Unfortunately, all these methods fail to accomplish the HPE tasks in the presence of occlusion where hands hidden behind, for example, cardboard or sleeves.

Recent research has demonstrated that it is possible to distinguish RF signals reflected by different parts of the human body~\cite{zhao2018through, wang2019person}. The shape and action amplitudes of body parts impact the intensity of reflected RF signals, thus enabling the reconstruction of human poses through deep analytics. However, these approaches are not directly applicable to tackle HPE, because they typically involve signal accumulation over time to detect body pose at a larger scale~\cite{zhao2018through, wang2019person}. Although RF has succeeded in addressing HGR~\cite{cai2019efficient, skaria2019hand}, these solutions are incompatible with what HPE demands for the same reason as explained earlier for CM-HGR. Again, none of these proposals is capable of handling HPE under occlusion.

\section{\name Design}  \label{sec:system} 
In this section, we present three key components of \name, namely a deep cross-modality framework, a deep complex-valued network \nname, as well as a deep adversarial learning algorithm \nnname.

\begin{figure*}[t]
	\centering
	\includegraphics[width=0.95\textwidth]{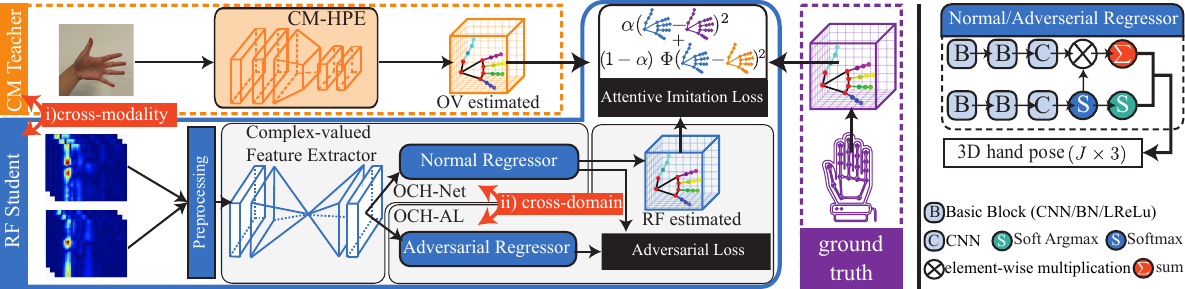} 
	\caption{Architecture of \name: i) cross modality from OV to RF and ii) cross-domain from LoS to occluded.}
	\label{fig:sys_overview}
     \vspace{-2ex}
\end{figure*}

\subsection{Overview} \label{ssec:overview} %
The overview of \name is illustrated in Figure~\ref{fig:sys_overview}. The camera and RF sensor are calibrated extrinsically according to their different positions~\cite{lim2021radical}, and synchronized using the network time protocol~\cite{mills1991internet}. Before feeding RF signals to the neural models, an RF preprocessing module is employed to suppress the noise and improve the quality of signals. The RF preprocessing module employs a smooth filter followed by a band-pass filter to process RF signals~\cite{chen2021rf}.
After data collection and preparation, \name consists of the following three major components:

\vspace{.5ex}
\noindent\textbf{C1:} We pre-train a CM-HPE network using ground truth obtained from OptiTrack and transfer knowledge cross-modality from CM-HPE to RF-HPE under LoS situation, by minimizing an attentive imitation loss (Section~\ref{ssec:dcm}), while the detailed construction of RF-HPE (feature extractor and regressor) is postponed to later sections.

\vspace{.5ex}
\noindent\textbf{C2:} To handle complexed-valued RF signals, we construct our RF-HPE network \nname as a complex-valued feature extractor and a 2D-3D regressor (Section~\ref{ssec:dcvn}).

\vspace{.5ex}
\noindent\textbf{C3:} For generalizing towards occluded domain with unseen obstacles, we further employ adversarial learning to transfer knowledge from the LoS domain to the occluded one. Specifically, an adversarial regressor is added to form a minimax game with the \nname regressor, allowing for an unsupervised training (Section~\ref{ssec:dal}). 

\vspace{.5ex}
Overall, the two-stage training of \name realizes both cross-modality from OV to RF and cross-domain from LoS to occluded, resulting in an occlusion-robust RF-HPE model. This is in stark contrast to the troublesome alternative that employs a synchronized camera at a vantage point (only if possible) for every occluded scenario to provide direct OV annotations and then trains an RF-vison model in a supervised manner.
Essentially, this alternative still involves cross-modality and cross-domain training that are not always feasible for all scenarios, hence it is hardly generalizable to unseen domains.

\subsection{Deep Cross-Modality Framework}  \label{ssec:dcm} 
In this section, we draw inspiration from recent research works~\cite{zhao2018through,wang2021knowledge, saputra2019distilling} and propose a teacher-student framework for cross-modality training, aiming to tackle the challenge of mapping RF signals to hand keypoints. 
For brevity, we use the notations $\mathsf{T}$ for the OV teacher network, $\mathsf{S}$ for the RF student network, and $\bm{y}^{\mathsf{T}}$, $\bm{y}^{\mathsf{S}}$, $\bm{y}$ for the predictions of the CM-HPE network, the RF-HPE network, and the ground truth, respectively. Unlike a classification task, we cannot utilize the ``dark knowledge'' of the soft output in the logit space for knowledge distillation of HPE regression.
Therefore, to effectively distill knowledge from the CM-HPE network we employ an attentive imitation loss~\cite{saputra2019distilling} as follows:
{\small
\begin{equation} \label{eq:lossy}
L_{y} = (1/n) \textstyle{\sum_{i=1}^n} \alpha(\bm{y}_i^{\mathsf{S}}-\bm{y}_i)^2+(1-\alpha) \Phi_i(\bm{y}_i^{\mathsf{S}}-\bm{y}_i^{\mathsf{T}})^2, \nonumber
\end{equation}
}
where $\alpha$ is a scale factor, $\Phi_i =\left(1-\frac{(\bm{y}_i^{\mathsf{T}}-\bm{y}_i)^2}{\eta}\right)$ is the normalized teacher loss for the $i$-th sample, $\eta =\max \left(e^{\mathsf{T}}\right)-\min \left(e^{\mathsf{T}}\right)$ is a normalization parameter obtained as the difference between
the maximum and minimum of $e^{\mathsf{T}}$, where $e^{\mathsf{T}} =\left\{(\bm{y}_j^{\mathsf{T}}-\bm{y}_j)^2: j=1, \ldots, N\right\}$ is a set of teacher loss from entire training data. In the loss function, $\Phi_i$ assigns different relative importance to each component, making the loss attentive. Since $\Phi_i$ is computed differently for each sample, it down-weights unreliable predictions made by the teacher $\mathsf{T}$. This means that, during knowledge transfer from $\mathsf{T}$ to $\mathsf{S}$, the cross-modality training relies more on the example data that $\mathsf{T}$ is good at predicting.

\subsection{\nname: A Deep Complex Perspective} \label{ssec:dcvn} %
Despite the guidance of the CM-HPE model, the RF-HPE model still faces challenges in understanding RF signals due to their complex-valued and high-dimensional tensor nature. Essentially, the \textit{RF tensor} consists of data from multiple antennas acquiring RF pulses reflected by targets in the environment. Since a single RF pulse is insufficient for vision, multiple pulses are transmitted both in time and space (via different antennas) and the received signals are combined to form the RF tensor $\bm X$, whose every element is a complex number with I/Q components $\bm{x}(n,t,d) = \bm{x}_{\text{I}} + j \bm{x}_{\text{Q}}$, where $n$, $t$ and $d$ respectively
indicate the transmitter-receiver pair in the antenna array, the number of RF pulses, and the number of discretized range bins~\cite{chen2021octopus}. In particular, the phase of each $\bm{x}(n,t,d)$ carries important information about the relative displacement of the targets. To fully exploit the potential in the I/Q components of RF data, we propose \nname that specializes in handling the complex-valued RF data tensor.
\begin{figure}[b]
    \vspace{-2ex}
	\centering
	\includegraphics[width=0.38\textwidth]{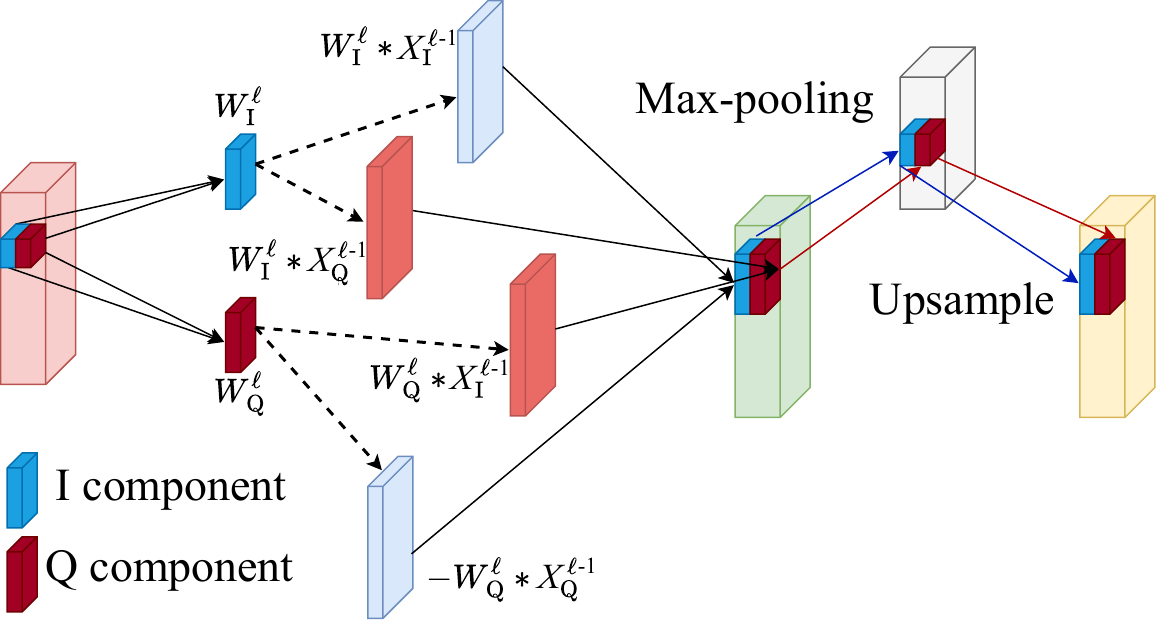}
        \vspace{1ex}
	\caption{Network operations for processing RF tensor.}
	\label{fig:cvnn_block}
    \vspace{-2ex}
\end{figure}

\vspace{-2ex}
\paragraph{Building Blocks for Complex} 
Before diving into the full architecture, we first introduce our updated version of 2D convolution building blocks for handling complex-valued RF tensor ${\bm X}=[\bm{x}(n,t,d)]$, in which $n$ is treated as the number of channels of 2D input,  $t$ and $d$ are treated as height and width. Conventional 2D CNN for real-valued data at the $\ell$-th layer could be described as $\bm{X}^{\ell} = \bm{W}^{\ell}  * \bm{X}^{\ell-1}$, where $\bm{W}^{\ell}$ is the convolution kernel and the symbol $*$ represents the convolution operator. 
Given $\bm{X} = \bm{X}_{\text{I}} + j \bm{X}_{\text{Q}}$ as the RF tensor, we perform the complex-valued convolution using two new real-valued kernels $\bm{W}_{\text{I}}^{\ell}$ and $\bm{W}_{\text{Q}}^{\ell}$~\cite{trabelsi2017deep}. 
Instead of simply stacking the two parts, we make $ \bm{X}^{\ell} =  ( \bm{W}_{\text{I}}^{\ell} * \bm{X}^{\ell-1}_{\text{I}}  - \bm{W}_{\text{Q}}^{\ell} * \bm{X}^{\ell-1}_{\text{Q}} ) + j ( \bm{W}_{\text{I}}^{\ell} * \bm{X}^{\ell-1}_{\text{Q}} + \bm{W}_{\text{Q}}^{\ell} * \bm{X}^{\ell-1}_{\text{I}} ) $. We also redefine the complex-valued nonlinear activated function as $\sigma_{\text{C}} ( \bm{X} ) = \sigma ( \bm{X}_{\text{I}} ) \oplus \sigma (\bm{X}_{\text{Q}}) $, where $\sigma$ is the original activation function, and $\oplus$ represents the operation of concatenation. The downsampling, upsampling, and batch normalization layers are similarly redefined
by processing the real and imaginary branches separately with their real-valued counterparts and then concatenating the results. This whole procedure is visualized in Figure~\ref{fig:cvnn_block}.

\vspace{-2ex}
\paragraph{\nname Architecture}
Leveraging the complex-valued building blocks, we design \nname with a feature extractor followed by a regressor, as illustrated in Figure~\ref{fig:sys_overview}. The feature extractor is based on a popular encoder-decoder architecture with skip connections~\cite{ronneberger2015u}. The encoder and decoder blocks are paired and connected via a skip connection to facilitate information flow. In this way, \nname utilizes fine-grained details learned in the encoder part to estimate hand poses in the decoder part. 

In particular, each encoder block in the system consists of a complex-valued convolutional layer, a batch normalization layer, and a nonlinear activation layer with leaky ReLU. 
After every three blocks, the number of channels increases, and a max-pooling layer is applied to enhance the most prominent feature and reduce the dimension of the hidden layers, thus reducing complexity. Once the bottleneck block extracts a representation in the latent space, an upsampling layer is utilized to reverse the compression for the decoding process. Similarly, each decoder block contains the same components as the encoder block. After every three blocks, the number of channels decreases, and an upsampling layer is inserted to maintain the shape of the feature map.  
Finally, the decoded features are fed to a regressor to map to 3D hand keypoints. As shown in the upper-right Figure~\ref{fig:sys_overview}, the lower branch of the regressor recovers the 2D hand keypoints, while the upper branch recovers the depth of the keypoints. Finally, the 3D keypoints are reconstructed by combining the 2D and depth estimations. 

\begin{figure}[t]
	\centering
	\includegraphics[width=0.42\textwidth]{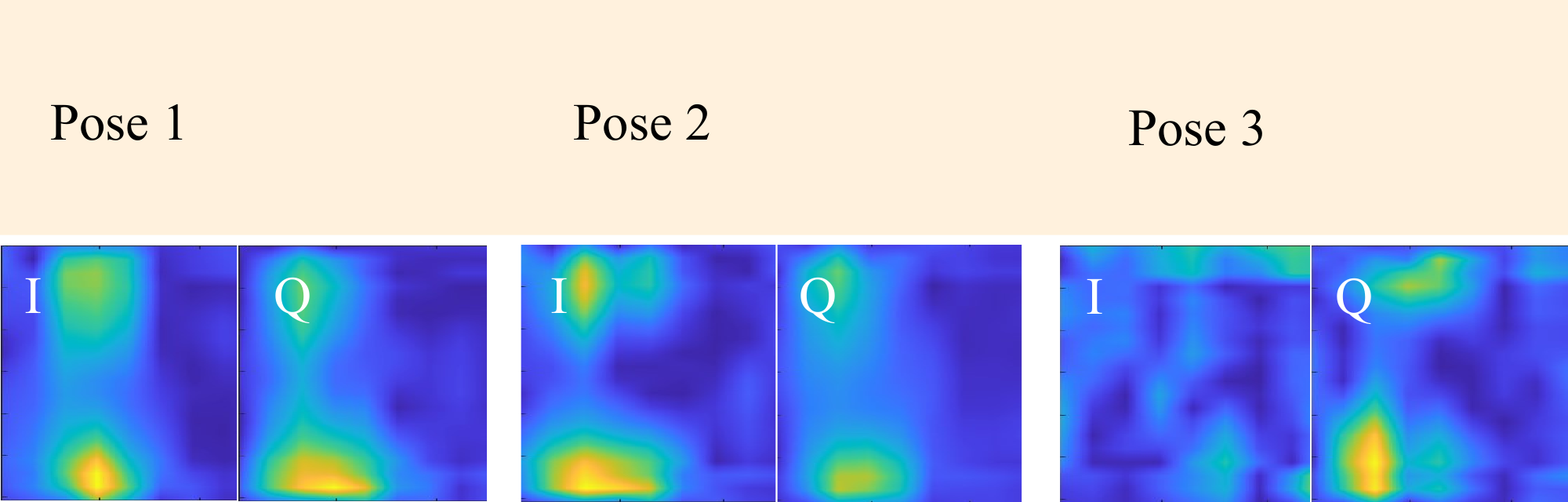}
        \vspace{1.5ex}
	\caption{Activated feature maps of different poses.} 
	\label{fig:rf_features}
     \vspace{-2ex}
\end{figure}

To demonstrate the necessity of using \nname to handle complex-valued RF signals, we visualize the feature maps in one of \nname's hidden layer for various hand pose RF data. The results in Figure~\ref{fig:rf_features} clearly indicate that I and Q data generates feature maps with distinctive differences, with bright yellow regions representing activated neurons in the feature maps. These findings highlight that relying solely on either I or Q component of the RF data as input can be insufficient for capturing all intrinsic features of RF data for estimating hand poses, potentially resulting in missing crucial information. 

\subsection{Deep Adversarial Learning for Occlusion} \label{ssec:dal}  %
Since \nname only works for the normal domain, we need to transfer the learned knowledge to the occluded domain.
We denote the $n$-th sample of the RF data as $\bm{x}_n$, and in unsupervised deep adversarial learning, we denote the labeled normal domain as $ D^{\text{NO}} = \{ (\bm{x}_n^{\text{NO}}, \bm{y}_n^{\text{NO}} ) \}_{n=1}^{N}$, and the unlabeled occluded domain as $ D^{\text{OC}} = \{  \bm{x}_m^{\text{OC}} \}_{m = 1}^{M}$. The objective of this training algorithm is to maximize the regressive accuracy in the occluded domain.

\vspace{-2ex}
\paragraph{Minimax Game} 
Due to the lack of OV annotations in $ D^{\text{OC}}  = \{  \bm{x}_m^{\text{OC}} \}_{m = 1}^{M}$, we need an unsupervised training approach. We introduce an adversarial regressor $g'$ to form a minimax game with the \nname regressor $g$, so as to minimize the expected loss of $g$ on the occluded domain while maintaining good performance on the normal domain. 
To this end, we leverage the disparity discrepancy theory~\cite{zhang2019bridging,jiang2021regressive} that allows for a proper alignment between two domains via disparity reduction.

We define the \textit{disparity} between two regressors $g$ and $g'$ as the expected loss over a domain $D$, denoted by $\operatorname{disp}_D\left(g', g\right) = \mathbb{E}_D L\left(g', g\right)$, and the \textit{disparity discrepancy} induced by $g'$ as the supremum of the difference between the disparities of $D^{\text{NO}}$ and $D^{\text{OC}}$ over a hypothesis space $\mathcal{G}$, denoted by $d_{g, \mathcal{G} }(D^{\text{NO}}, D^{\text{OC}}) = \sup_{{ {g' \in \mathcal{G}}}}  ( \operatorname{disp}_{D^{\text{OC}}} (g', g) - \operatorname{disp}_{D^{\text{NO}}} (g', g) )$. 
It is proven in~\cite{zhang2019bridging} that we can strictly bound the expected error of $g$ on the occluded domain by the following minimization objective: 
{\small
\begin{align} 
	\min_{g \in \mathcal{G}} \epsilon_{D^{\text{NO}}} (g) +  d_{g, \mathcal{G} }(D^{\text{NO}}, D^{\text{OC}}), \nonumber
\end{align}
}
where $\epsilon_{D^{\text{NO}}} (g) = \mathbb{E}_{(\bm{x}^{\text{NO}}, \bm{y}^{\text{NO}}) \in {D^{\text{NO}}} }  L_a (g(\bm{x}^{\text{NO}}), \bm{y}^{\text{NO}} )$ is the expected regression loss $L_a$ of $g$ in the normal domain. Since $\epsilon_{D^{\text{NO}}} (g)$ is determined by $g$ during the pre-training stage, only the second term, i.e., the disparity discrepancy between the two domains, needs to be minimized. 
This term can be approximated by maximizing over $g'$ as a deep learning model instead of taking supremum in the space $\mathcal{G}$, with $\phi$ as the parameter-fixed feature extractor:
{\small
\begin{align}  \label{eq:ad_reg}
d_{g, \mathcal{G} }&(D^{\text{NO}}, D^{\text{OC}})\approx  \max_{g'} \left( \operatorname{disp}_{D^{\text{OC}}} (g', g) - \operatorname{disp}_{D^{\text{NO}}} (g', g) \right)\nonumber \\
  = &\max_{g'}  \mathbb{E}_{ \bm{x}^{\text{OC}}\in D ^{\text{OC}}  } L_a \left( ( g'  \circ \phi )(\bm{x}^{\text{OC}}),   ( g  \circ \phi )(\bm{x}^{\text{OC}})  \right)  \nonumber \\
	- &\mathbb{E}_{ \bm{x}^{\text{NO}}\in D^{\text{NO}} } L_a \left( ( g'  \circ \phi )(\bm{x}^{\text{NO}}),   ( g  \circ \phi )(\bm{x}^{\text{NO}}) \right). 
\end{align}
}
After training $g'$ to approximate the disparity discrepancy $d_{g, \mathcal{G} }(*)$, minimizing the following equation will decrease the error of $g$ in the occluded domain effectively:
{\small
\begin{equation}  
	\min_{\phi, g}  \mathbb{E}_{ (\bm{x}^{\text{NO}}, \bm{y}^{\text{NO}} ) \in {D^{\text{NO}}}  } L_a \left( ( g  \circ \phi )(\bm{x}^{\text{NO}}),  \bm{y}^{\text{NO}} \right) + d_{g, \mathcal{G} }(*). \nonumber
\end{equation}
}
To implement this step, we fix the parameters of $g'$, and update the parameters of $g$ and $\phi$ by backpropagation. This process essentially creates a minimax game between two regressors, each working towards achieving opposing goals. However, their collaboration enables the adaptation from the normal domain to the occluded domain. 

\vspace{-2ex}
\paragraph{Bounding Output Space.} 
Compared with a classification task, HPE involves regression with a much larger output space. Specifically, if we consider a target hand pose existing in a 3D voxel space with dimensions of height $H$, width $W$, and depth $R$, and treat each output voxel as a class, we would have $H \times W \times R$ classes. Consequently, the large number of output classes will increase the bound of occluded domain error. Therefore, it is necessary to reduce the output space for the HPE network. As pointed out by~\cite{jiang2021regressive}, there is an intrinsic sparsity in the keypoint positions: when inferring hand poses under occlusion using $g \circ \phi$, even if the estimated hand pose is incorrect, the keypoint positions are still on the hand and likely overlap with other keypoint positions. For instance, the incorrect little finger keypoint position may appear on the index finger's position, but not in the background (more examples are illustrated in Section~\ref{sssec:al_results}). This indicates that the output space size can be bounded to a limited set.

Therefore, during the training of \nnname, it only needs to pay more attention to the limited set. To achieve this, we can accumulate all incorrectly predicted hand poses while training in the normal domain and compute the distribution of each false hand pose heatmap ${m}_{\text{GF}}( \hat{ \bm{y}}_i^{\text{NO}}) = \sum_{q \ne i} \mathcal{N}( \hat{\bm{y}}_q^{\text{NO}})$ where $\mathcal{N}$ is the 3D truncated Gaussian function, and $\hat{ \bm{y}_i^{\text{NO}} }$ is the $i$-th keypoint heatmap of a hand pose predicted by the normal domain regressor $g$. Furthermore, in Eqn.~\eqref{eq:ad_reg}, we observe that only the occluded predictions $( g'  \circ \phi )(\bm{x}^{\text{OC}})$ are used to maximize the disparity $ \mathbb{E}_{ \bm{x}^{\text{OC}}\in {D^{\text{OC}}}  } L_a ( ( g'  \circ \phi )(\bm{x}^{\text{OC}}),   ( g  \circ \phi )(\bm{x}^{\text{OC}})  )$ by updating the adversarial regressor $g'$. Therefore, since the estimated supremum of the disparity discrepancy likely occurs on a false prediction, we can further leverage the false heatmap distribution ${m}_{\text{GF}}( \hat{ \bm{y}}_i^{\text{NO}})$ to get $\tilde{ \mathcal{L}}_{\text{a}}  ( \bm{x}^{\text{OC}})  = \mathbb{E}_{ \bm{x}^{\text{OC}}\in {D^{\text{OC}}}  } L_a (  ( {m}_{\text{GF}} \circ g'  \circ \phi )(\bm{x}^{\text{OC}}) ),   ( g \circ \phi )(\bm{x}^{\text{OC}})  )$ , and revise  Eqn.~\eqref{eq:ad_reg} with the updated first term to obtain:
{\small
\begin{align}  
	\textstyle{\max_{g'}} d_{g, \mathcal{G} }(D^{\text{NO}}, D^{\text{OC}}) &=  \tilde{ \mathcal{L}}_{\text{a}} ( \bm{x}^{\text{OC}}) -  \mathcal{L}_{\text{a}}( \bm{x}^{\text{NO}}), \nonumber
\end{align}
}
where $ \mathcal{L}_{\text{a}}( \bm{x}^{\text{NO}})  =  \mathbb{E}_{ \bm{x}^{\text{NO}}\in {A}  } L_a ( ( g'  \circ \phi )(\bm{x}^{\text{NO}}),   ( g  \circ \phi )(\bm{x}^{\text{NO}}) )$.

\section{Evaluation} \label{sec:eval} %

\subsection{Experiment Setup}

\paragraph{Data Collection} 
For OV dataset,
we use a camera with $1080 \times 1920$ pixels and 30~\!Hz frame rate. For RF dataset,
since accessing raw signals from commodity devices (e.g., iPhone) is impossible, we emulate such an RF sensor by an IR-UWB radar~\cite{xethru} with 10 antennas~\cite{chen2021octopus}. The frame rate of the RF sensor is set to 150~\!Hz, and we use a Rockchip PX30~\cite{px30} to control the sensor. Both the camera and RF sensor are connected to a PC to be synchronized.
OptiTrack~\cite{optitrack} is used for obtaining 3D hand pose ground truth. 

We collect data with 30 volunteers in 5 different environments including a classroom, a living room, a bedroom, a lab cubicle, and a conference room. \camrev{We use hand poses from American Sign Language~\cite{liddell1989american} (including transition hand poses between two signs) along with randomly moving wrists and fingers, covering nearly all feasible variations of the hand's degrees of freedom.} The dataset includes 20 hours of normal condition data and 20 hours of occluded data. In the occluded scenarios, we utilize a variety of obstacles including wood, plastic, (frosted) glass, and cardboard sheets. These sheets are available in different areas, ranging from $0.5\text{m}^2$ to $1\text{m}^2$, with various widths between 1~\!cm and 10~\!cm, and placed at 10~\!cm in front of the RF sensor. The distance from the hand to the RF sensor varies from 20~\!cm to 80~\!cm. To annotate the OV dataset, we attach motion capture markers to the keypoints on a hand and retrieve the corresponding 3D coordinates from OptiTrack. Our evaluations use 10 hours of normal condition data (120,000 samples) to train \nname, then these data with additional 2 hours of occluded data (24,000 samples) to train \nnname, and finally all the remaining data are used for testing. 

\vspace{-2ex}
\paragraph{Teacher Network} 
We use MediaPipe Hands~\cite{zhang2020mediapipe}, a widely adopted CM-HPE network in many applications as our teacher network. We specifically set two key parameters of MediaPipe Hands, i.e., the maximum number of hands and minimum detection confidence as 1 and 0.5, respectively. The network is first pre-trained with OV data and ground truth collected by OptiTrack. Subsequently, we leverage MediaPipe's output and the ground truth to transfer knowledge to \nname.

\vspace{-2ex}
\paragraph{Training Details} 
We train and evaluate our method on a server with NVIDIA RTX 1080 GPU. We implement the \nname and \nnname on Python 3.8.16 and Pytorch 1.10.0. 
The input RF tensors have the size of $10\times40\times40$, with the scale factor $\alpha$ set to 0.5. In the feature extractor, the number of channels is set to 10, 64, 128, 256, and 512 in the encoder, and 512, 256, 128, 64, and 32 in the decoder for each convolutional layer. The regressor, as mentioned in Section~\ref{ssec:dcvn}, takes in concatenated real and imaginary features of 32 channels each to have in total 64 channels. These channels are then used to predict 21 keypoint heatmaps. We set the batch size to 8, and adopt an Adam optimizer with a learning rate of 0.001, $\beta_1$ of 0.9, and $\beta_2$ of 0.999. 

\vspace{-2ex}
\paragraph{Evaluation Metric}
The performance of HPE is evaluated using the percentage of correct keypoint~(PCK) metric~\cite{moon2020interhand2, spurr2018cross, spurr2021self} defined as follows: 
{\small
\begin{align}  \label{eq:pck}
	\text{PCK}@a = \frac{1}{N} \sum_{n = 1}^{N} \Xi \left( \frac{ \|  \bm{y}^{\text{pred}}_{n} - \bm{y}^{\text{gt}}_{n} \|^2_2 }{  \sqrt{ w_n^2  + h_n^2 + d_n^2} }  \le a \right), \nonumber
\end{align}
}
where $N$ is the number of test samples, $\Xi$ is a logical operation that outputs 0 if the expression is false and 1 if true,  $\bm{y}^{\text{pred}}_{n}$ denotes the predicted keypoint position, $\bm{y}^{\text{gt}}_{n}$ denotes the ground truth keypoint position, and $ \sqrt{ w_n^2  + h_n^2 + d_n^2}$ is the bounding box size of the hand. The PCK score ranges from 0 to 1, with higher scores indicating better performance. Typically, a normalized distance error of $a=0.2$ is used as the threshold for successful HPE~\cite{moon2020interhand2, spurr2018cross, spurr2021self}. To gain a clearer understanding of the performance of different parts of the hand, we also calculate PCKs at metacarpophalangeal  (MCP), proximal interphalangeal (PIP), distal interphalangeal (DIP), and fingertip joints, as defined in~\cite{spurr2021self}.

\subsection{Performance Evaluations} \label{ssec:results}

\subsubsection{Performance of \nname} 
We study the overall performance of \nname~\camrev{using normal condition data }with two existing RF vision methods as baselines, namely \textit{Person-in-WiFi}~\cite{wang2019person} and \textit{RF-Pose}~\cite{zhao2018through}. To the best of our knowledge, there is no directly related research work designed for hand pose with RF vision. Therefore, we have to modify the current human body skeleton neural network models~\cite{wang2019person,zhao2018through} to make them comparable to \nname. To be specific, we maintain the main part of their network architecture but replace their input and output to match our HPE task. Moreover, \nname can adapt to a different number of RF data streams as the network considers it as the number of input channels. To validate this, we create another baseline \textit{\nname-Slim} by extracting 2 data streams out of the 10 data streams collected from the RF antennas. In this experiment, we test all approaches under both normal and occluded scenarios.

\begin{figure}[t]
    \vspace{-1.5ex}
	\centering
	\subfloat[MCP. ]{ 
		\begin{minipage}[b]{.48\linewidth}
			\centering
			\includegraphics[width = \textwidth]{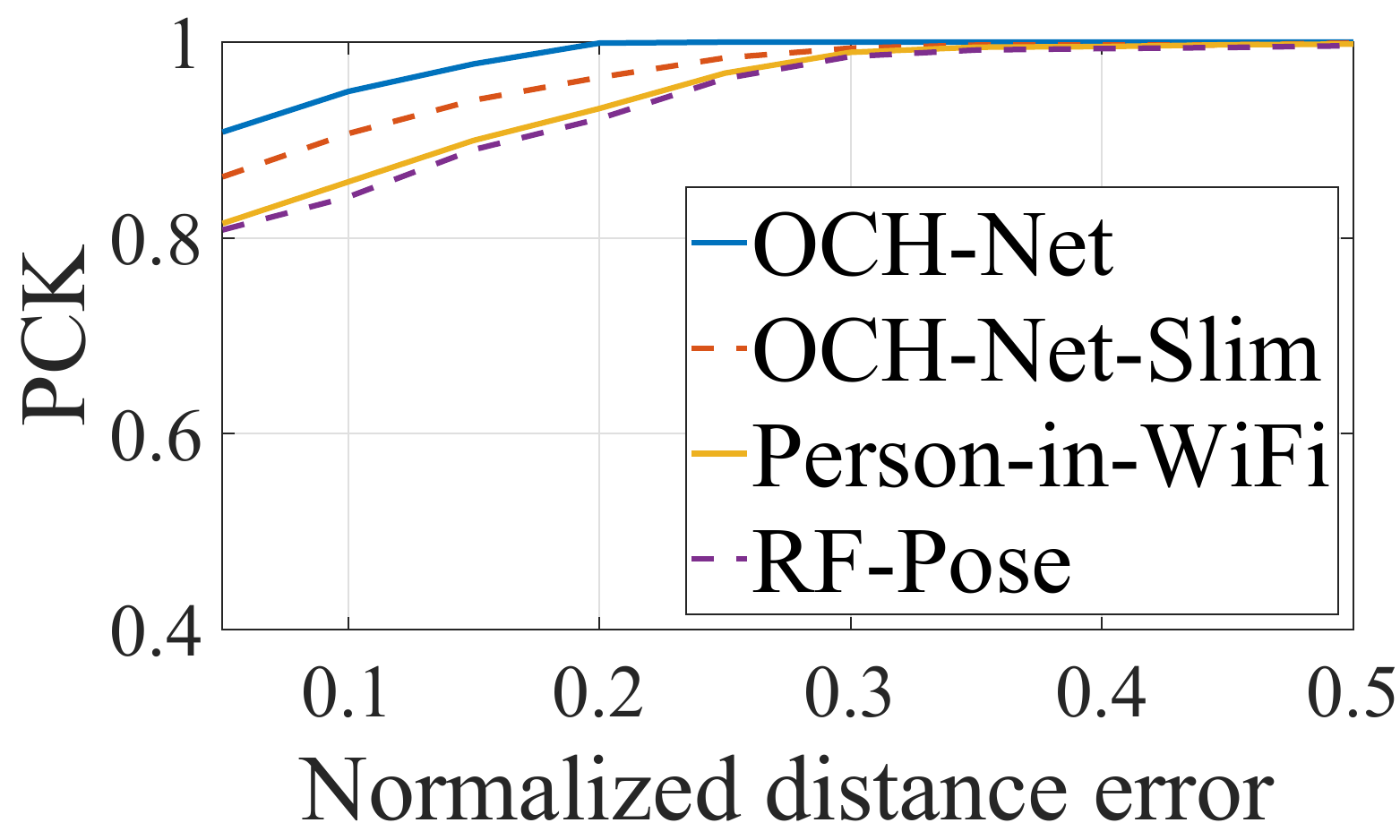}
                \vspace{-2.5ex}
			\label{sfig:op_mcp}
		\end{minipage}
	}
	\subfloat[{PIP.}]{ 
		\begin{minipage}[b]{.48\linewidth}
			\centering
			\includegraphics[width = \textwidth]{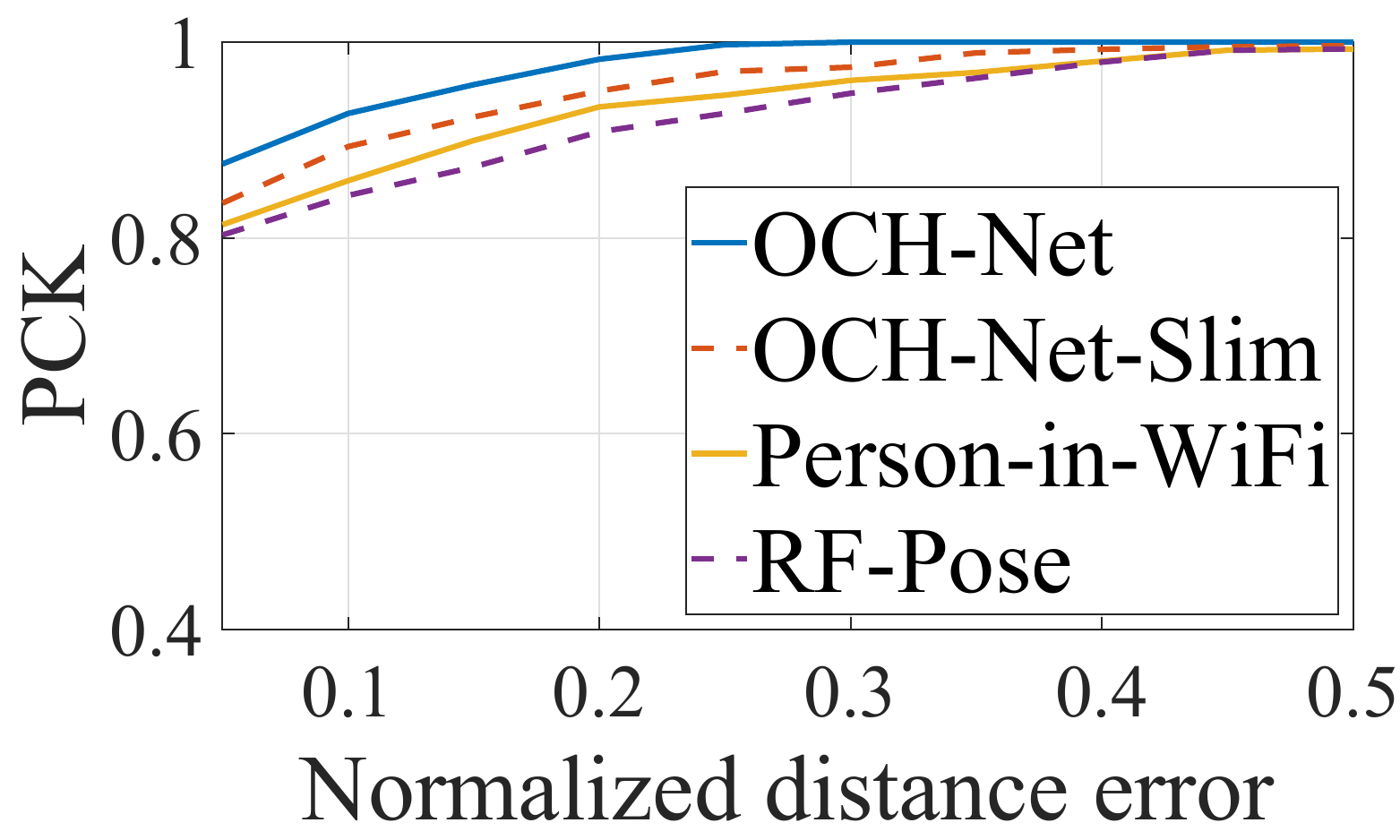}
                \vspace{-2.5ex}
			\label{sfig:op_pip}
		\end{minipage}
	}
        \\ \vspace{-2ex}
	\subfloat[DIP.]{ 
		\begin{minipage}[b]{.48\linewidth}
			\centering
			\includegraphics[width = \textwidth]{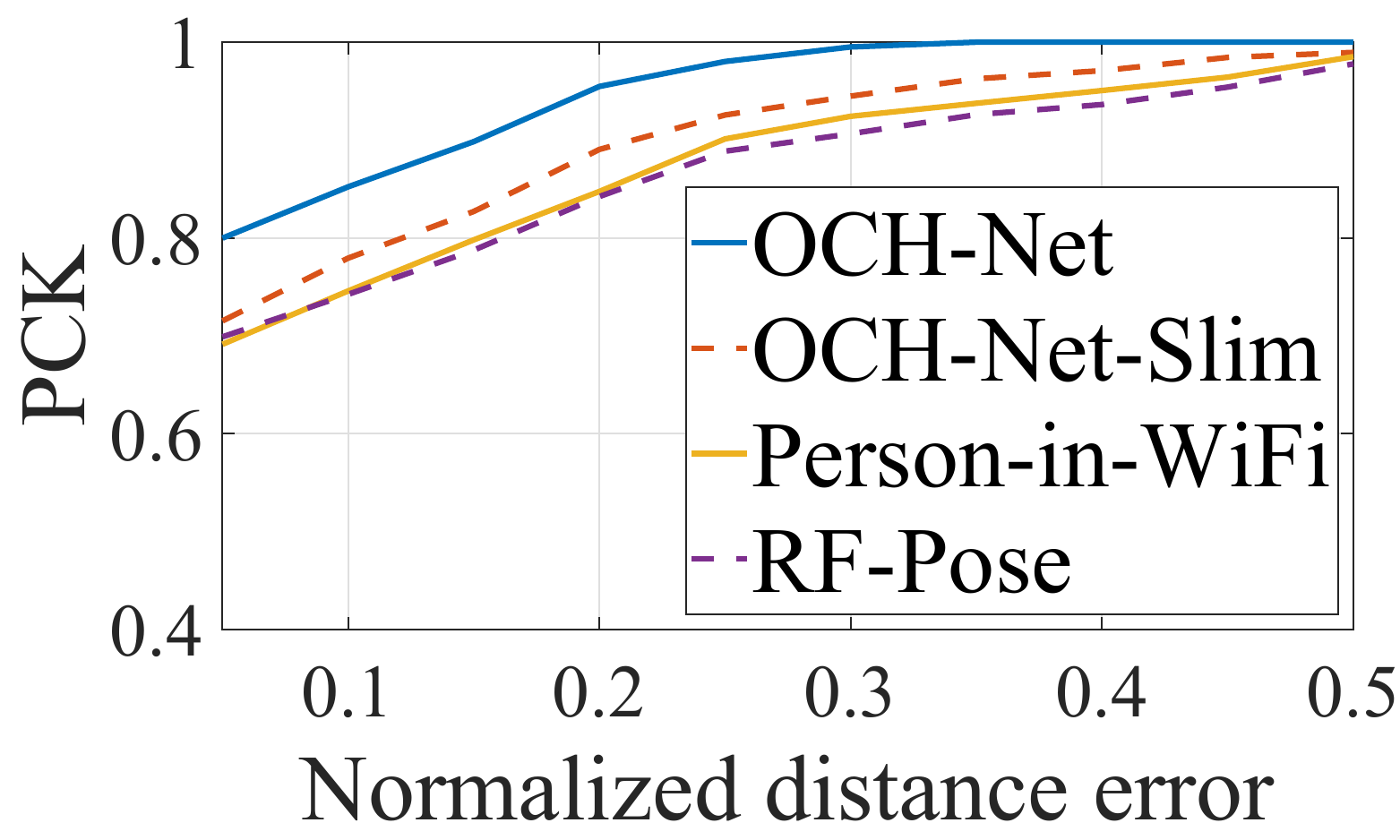}
                \vspace{-2.5ex}
			\label{sfig:op_dip}
		\end{minipage}
	}
	\subfloat[Fingertip.]{ 
		\begin{minipage}[b]{.48\linewidth}
			\centering
			\includegraphics[width = \textwidth]{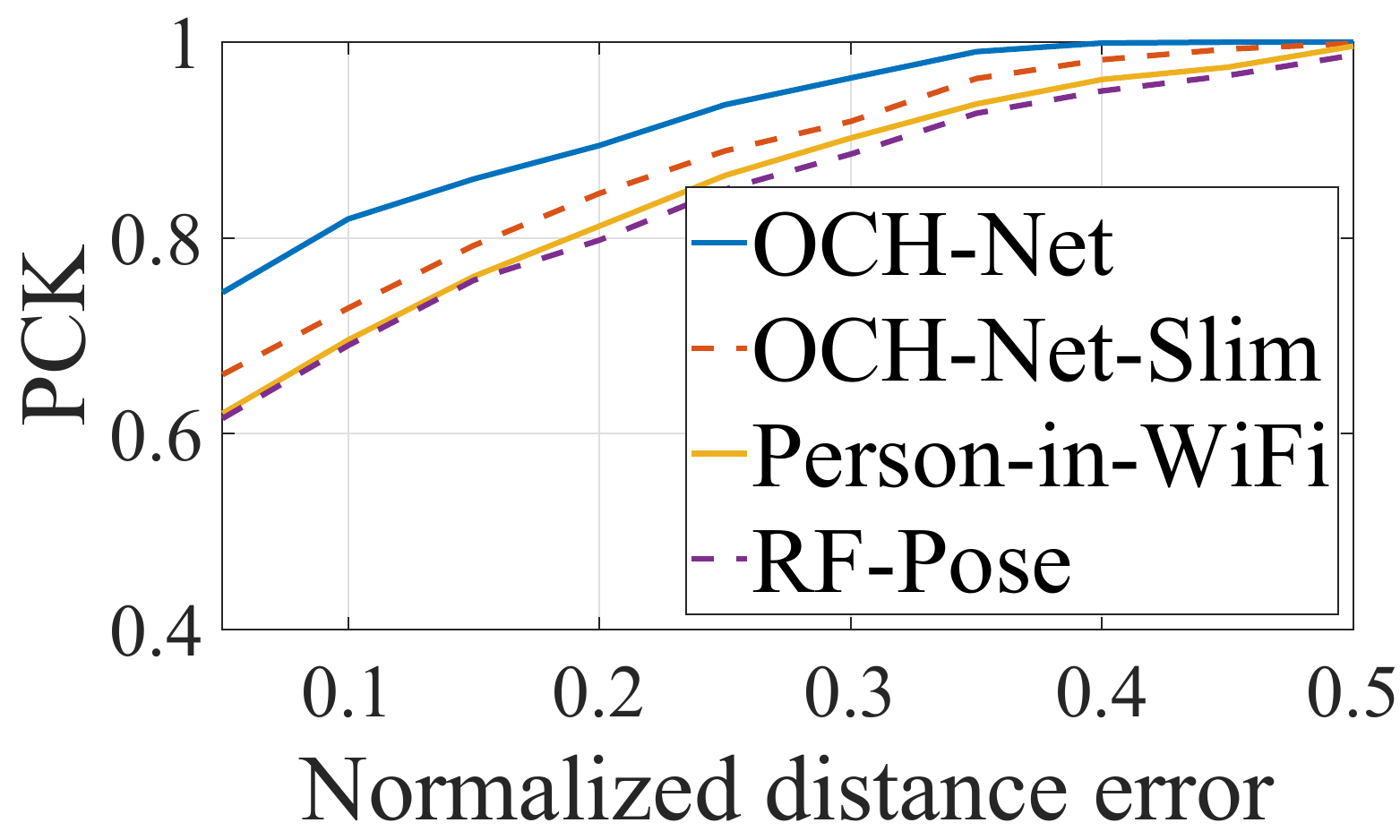}
                \vspace{-2.5ex}
			\label{sfig:op_fingertip}
		\end{minipage}
	}
        \vspace{1.5ex}
	\caption{The PCKs of \nname and different baselines.}
	\label{fig:finger_op}
        \vspace{-4ex}
\end{figure}

The results of the experiment, as shown in Figure~\ref{fig:finger_op}, demonstrate that both \nname and \nname-Slim outperform Person-in-WiFi and RF-Pose in terms of PCK@0.2 for all parts of the hand. In particular, \nname outperforms the two baselines by more than 10\%. Furthermore, since \nname utilizes more RF data streams, it also outperforms \nname-Slim by a small margin of approximately 5\% for all parts of the hand.  
There are two main advantages of using \nname over the two baselines. On one hand, \nname gains its knowledge from both the ground truth and the teacher network, effectively bridging the gap between complex RF-vision data and hand keypoints. This enables us to achieve non-Euclidean mapping. On the other hand, the deep complex-valued building blocks in \nname are better suited for interpreting and encoding RF data, as explained in Section~\ref{ssec:dcvn}. 

\begin{figure}[b]
 \vspace{-4ex}
        \captionsetup[subfigure]{justification=centering}
		\centering
	    \subfloat[3D HPE before and after \nnname under LoS and occlusion.]{
		  \begin{minipage}[h]{.92\linewidth}
		        \centering
			    \includegraphics[width = \textwidth]{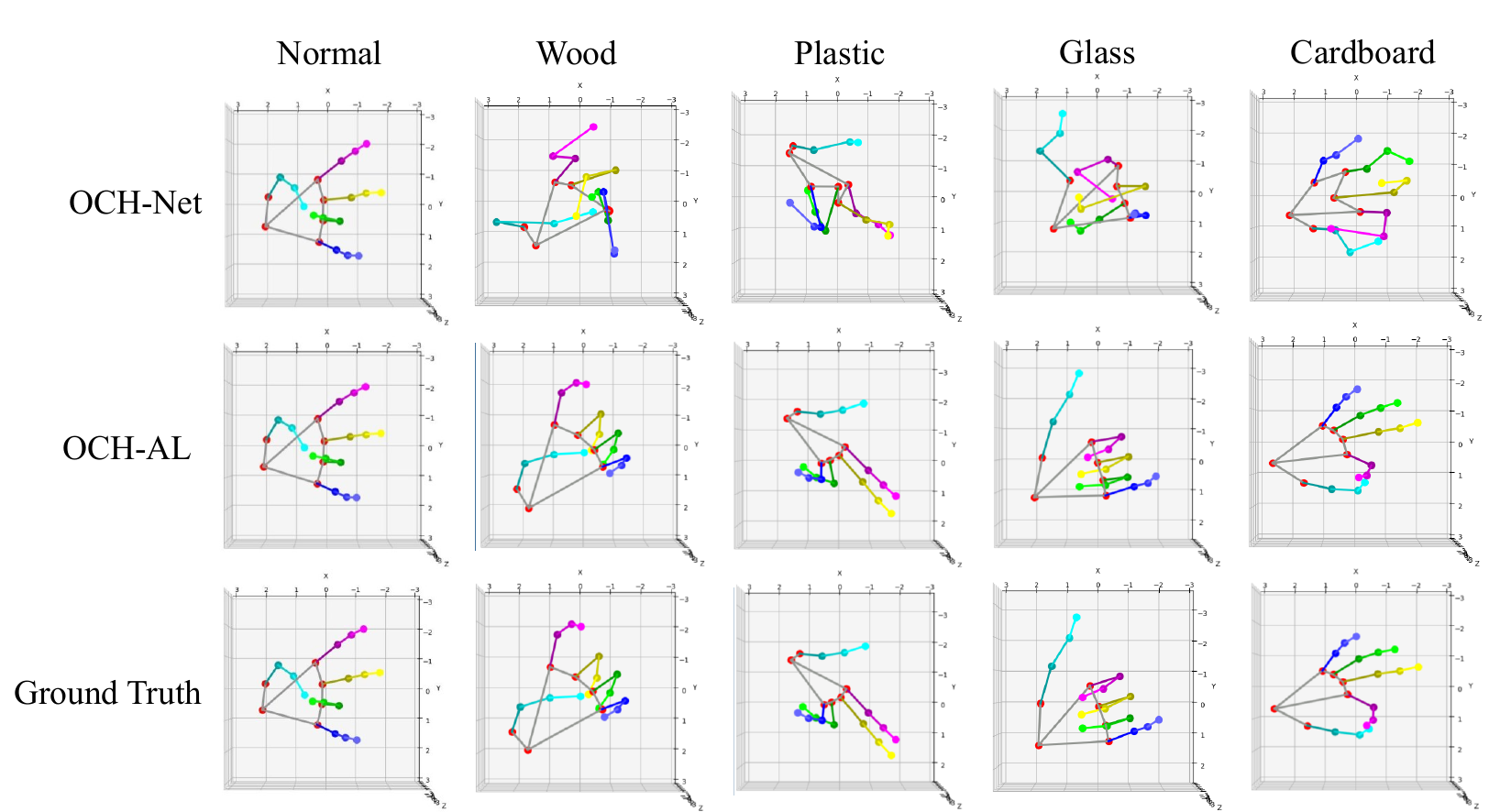}
                    \vspace{-3ex}
			    \label{subfig:qualitative_result_pose}
			\end{minipage}
		}
		\\
		\subfloat[Feature maps of before and after \nnname under occlusion.]{
		  \begin{minipage}[h]{.92\linewidth}
		        \centering
			    \includegraphics[width =\textwidth]{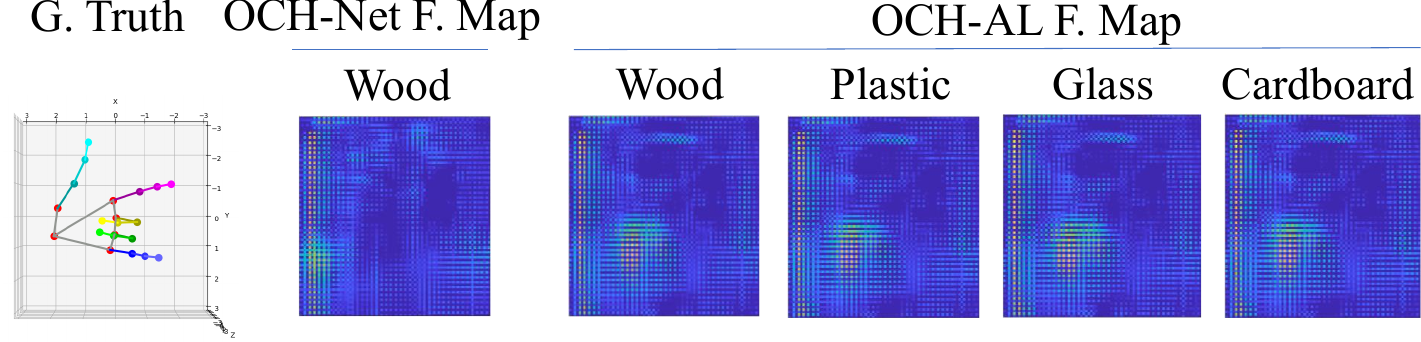}
                    \vspace{-1.5ex}
		      \label{subfig:qualitative_result_map}
			\end{minipage}
		}
        \vspace{1.5ex}
        \caption{Qualitative results before and after \nnname.}
	\label{fig:qualitative_result}
\end{figure}

\vspace{-2ex}
\subsubsection{Performance of \nnname under Occlusion}\label{sssec:al_results}
As described in Section~\ref{ssec:dal}, we employ the \nnname framework to adapt from the normal domain into the occluded one. Data from all 4 types of obstacles, namely wood, plastic, glass, and cardboard sheets are used for adaptation. To demonstrate the performance of \nnname, we present examples of recovered poses under various occluded scenarios in Figure~\ref{subfig:qualitative_result_pose}: while \nname can partially recover hand poses under occlusion, most of the recovered poses deviate from the ground truth, with a few joints overlapping with each other. In comparison, \nnname successfully adapts to the occluded domain and learns to handle the ``twisted'' RF signals caused by occlusion. We also show the feature maps at the bottleneck of the feature extractor in Figure~\ref{subfig:qualitative_result_map}; it further confirms that \nname achieves almost invariant feature maps for the same gesture despite occluded by distinct obstacles.

\begin{figure}[t]
    \vspace{-2ex}
	\centering
	\subfloat[MCP. ]{ 
		\begin{minipage}[b]{.48\linewidth}
			\centering
			\includegraphics[width = \textwidth]{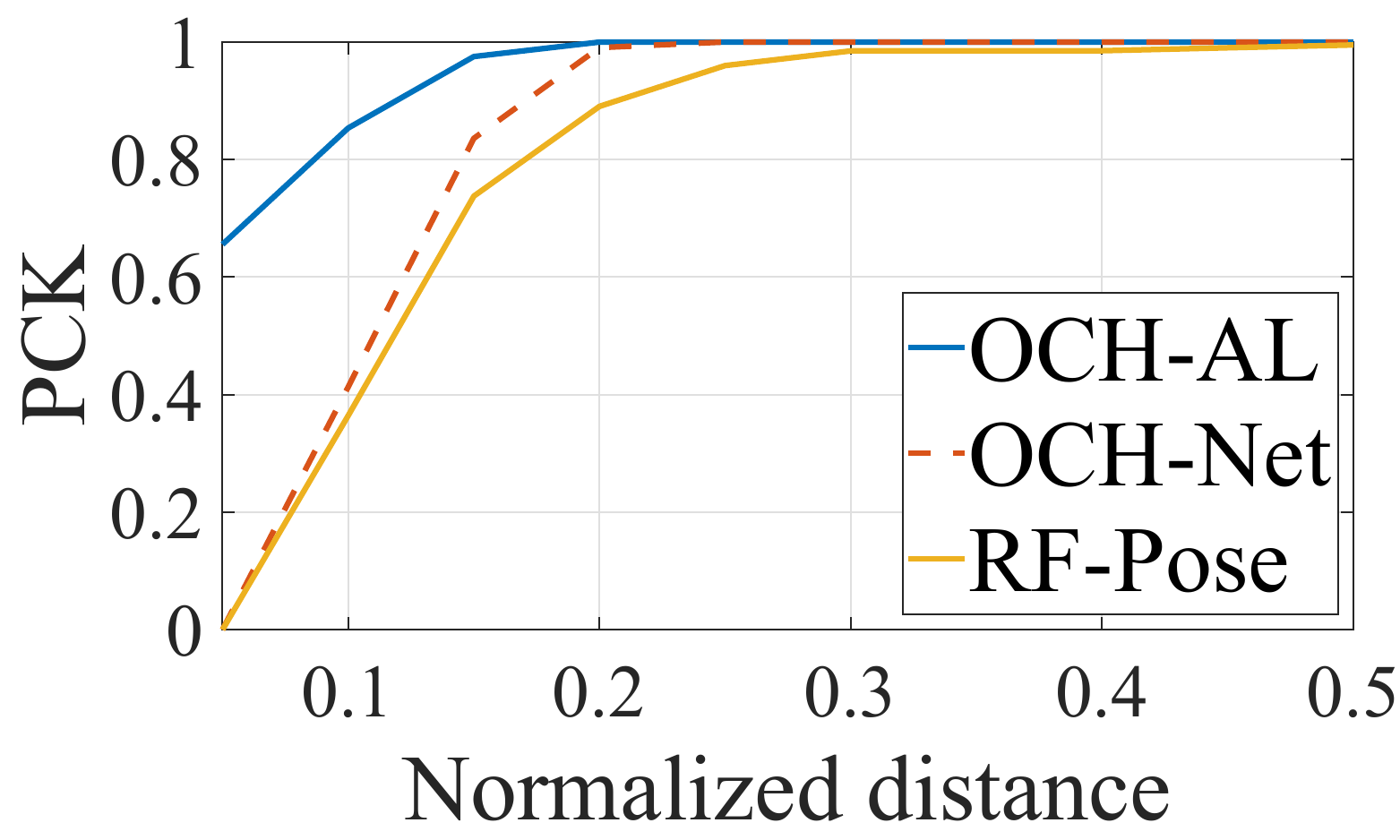}
                \vspace{-2.5ex}
			\label{sfig:ad_mcp}
		\end{minipage}
	}
	\subfloat[{PIP.}]{ 
		\begin{minipage}[b]{.48\linewidth}
			\centering
			\includegraphics[width = \textwidth]{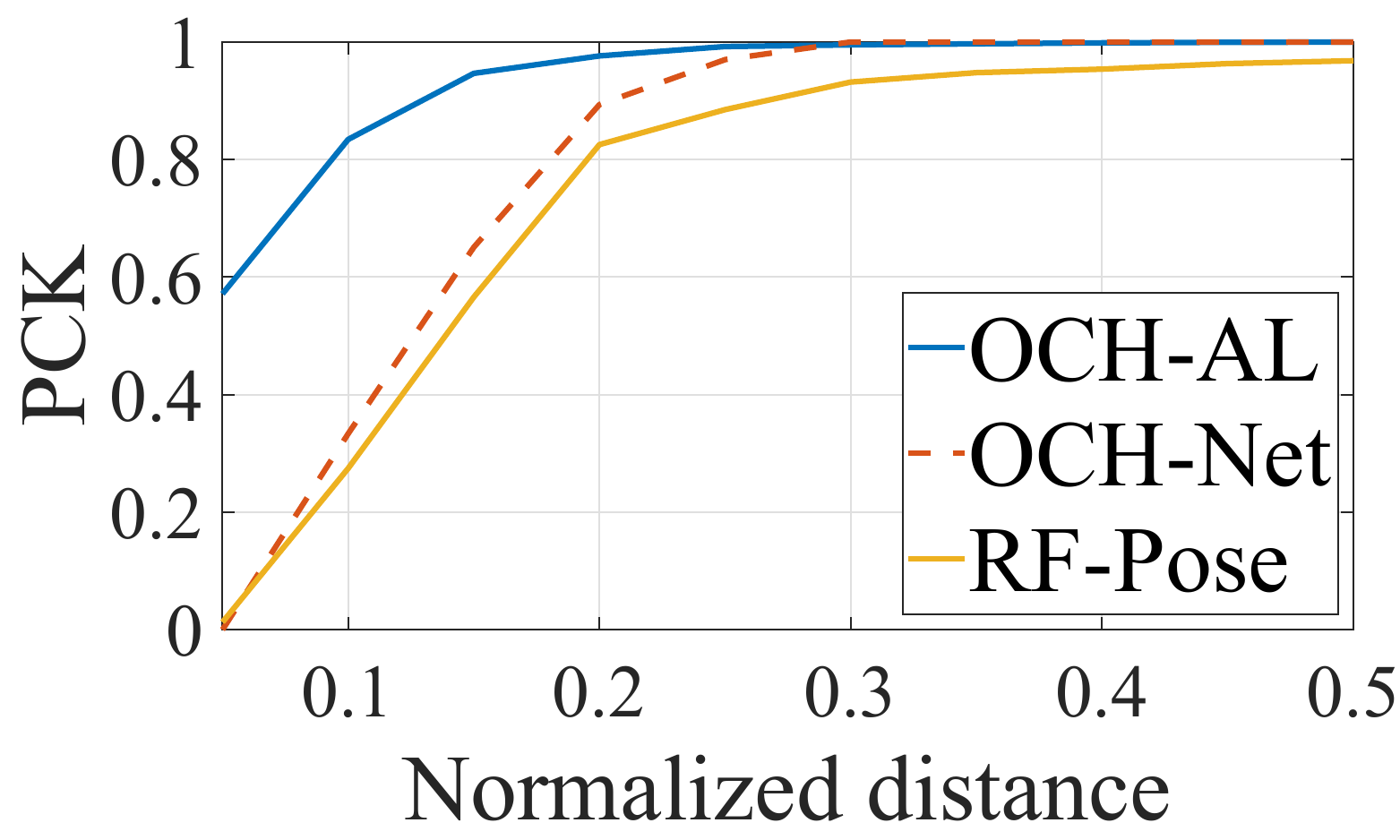}
                \vspace{-2.5ex}
			\label{sfig:ad_pip}
		\end{minipage}
	} 
        \\ \vspace{-2ex}
	\subfloat[DIP.]{ 
		\begin{minipage}[b]{.48\linewidth}
			\centering
			\includegraphics[width = \textwidth]{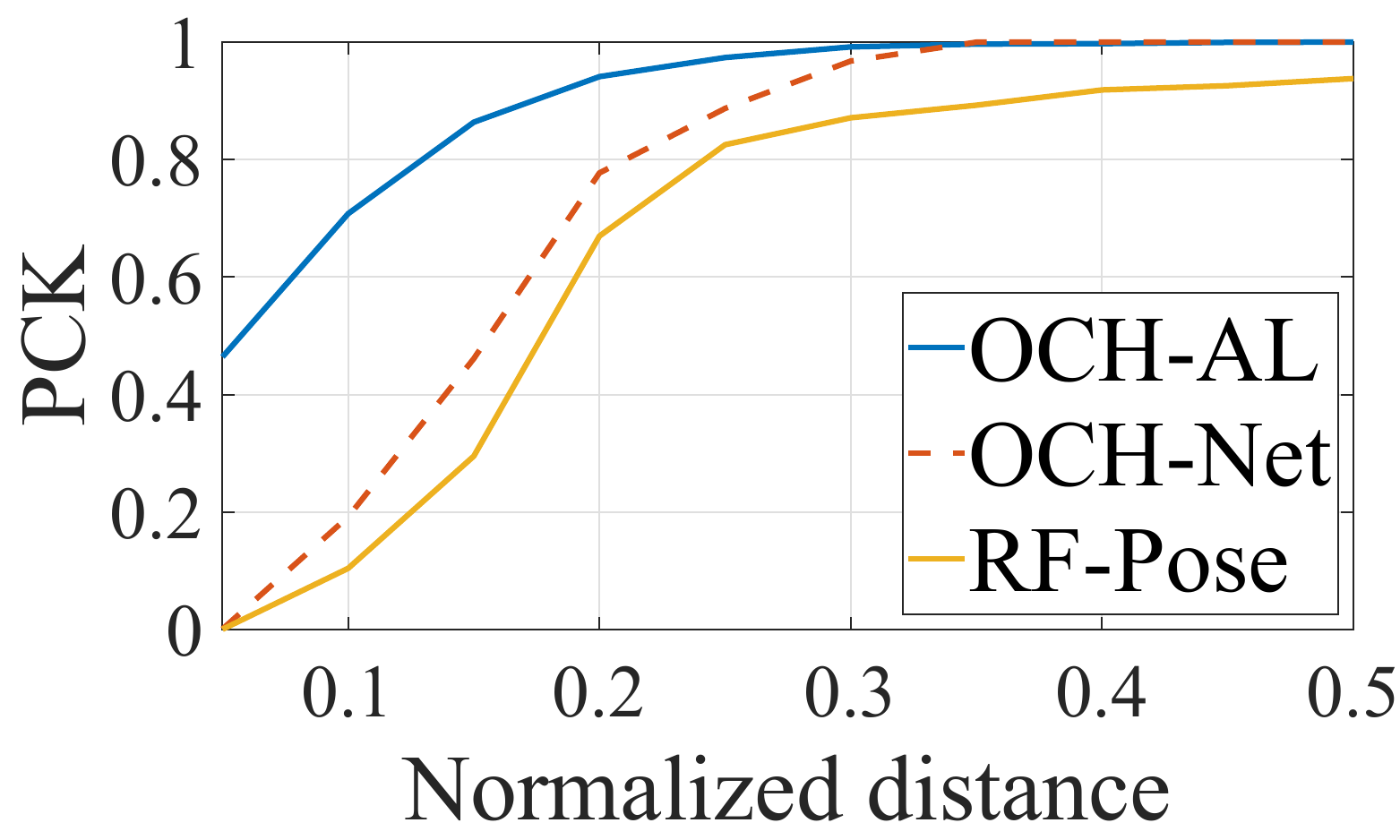}
                \vspace{-2.5ex}
			\label{sfig:ad_dip}
		\end{minipage}
	}
	\subfloat[Fingertip.]{ 
		\begin{minipage}[b]{.48\linewidth}
			\centering
			\includegraphics[width = \textwidth]{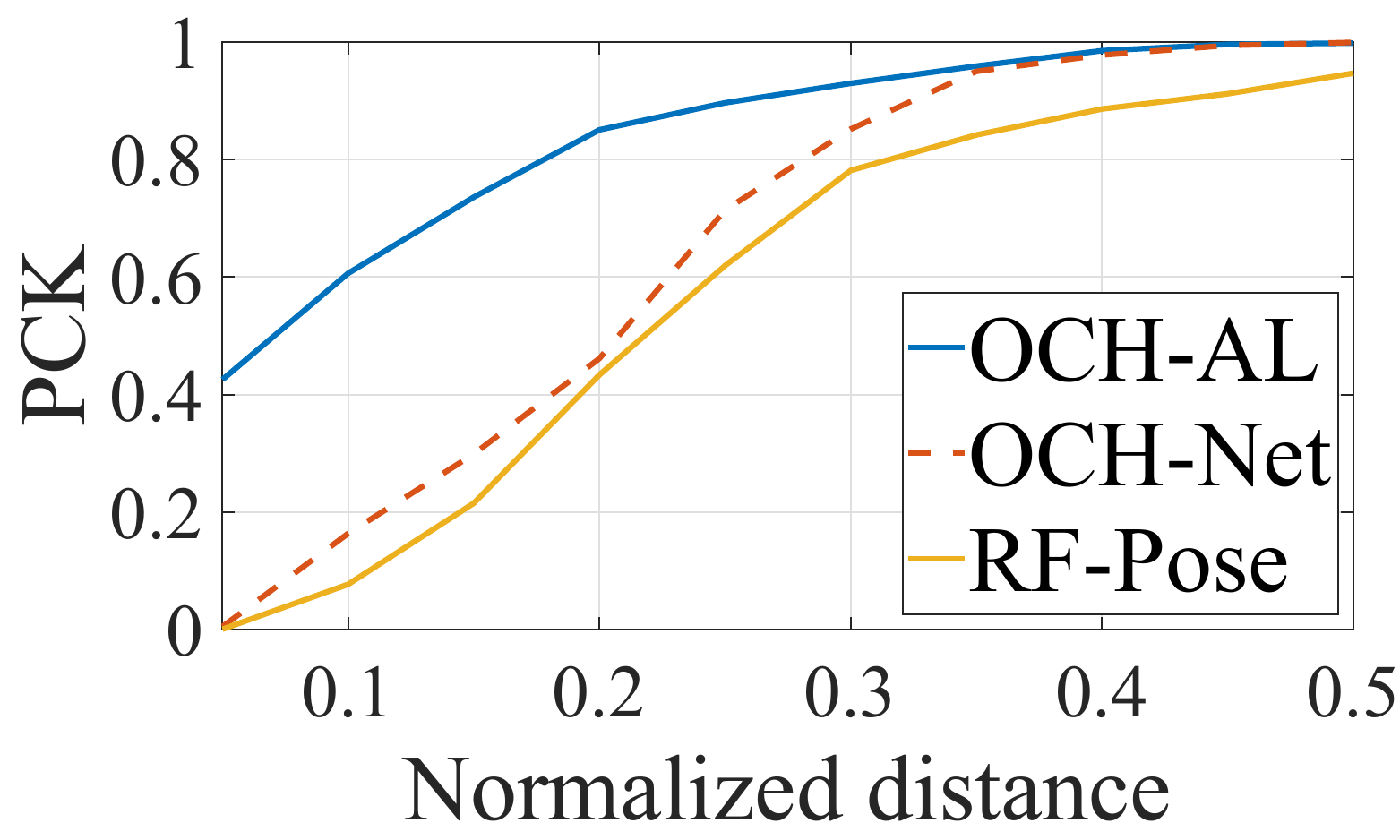}
                \vspace{-2.5ex}
			\label{sfig:ad_fingertip}
		\end{minipage}
	}
        \vspace{1.5ex}
	\caption{PCKs of \nname and \nnname under occlusion.}
	\label{fig:finger_ad}
    \vspace{-2ex}
\end{figure}

We further plot the performance of \nnname for different keypoints in Figure~\ref{fig:finger_ad}. The results show \nnname achieves PCKs@0.2 of 0.9998, 0.9763, 0.9410, and 0.8506 for MCP, PIP, DIP, and fingertip, respectively. In comparison, the PCKs@0.2 of \nname for the same keypoints are 0.9904, 0.8934, 0.7772, and 0.4615, respectively. These findings demonstrate \nnname successfully adapts \nname from the normal domain to the occluded one, resulting in significant performance improvements for all hand parts. \rev{Moreover, \nnname outperforms RF-Pose without domain adaptation, in terms of PCKs@0.2, by 0.1094, 0.1510, 0.2713, and 0.4169 for MCP, PIP, DIP, and fingertip respectively.} These improvements emphasize the need for \nnname: although RF-vision can bypass obstacles, its signals may be substantially altered by obstacle materials. Such signal variations result in different data distributions, rendering domain adaptation a necessary step.

\vspace{-2ex}
\subsubsection{Generalization to Unseen Occluded Scenarios}
\begin{figure}[h]
    \vspace{-1ex}
	\centering
	\subfloat[MCP. ]{ 
		\begin{minipage}[b]{.48\linewidth}
			\centering
			\includegraphics[width = \textwidth]{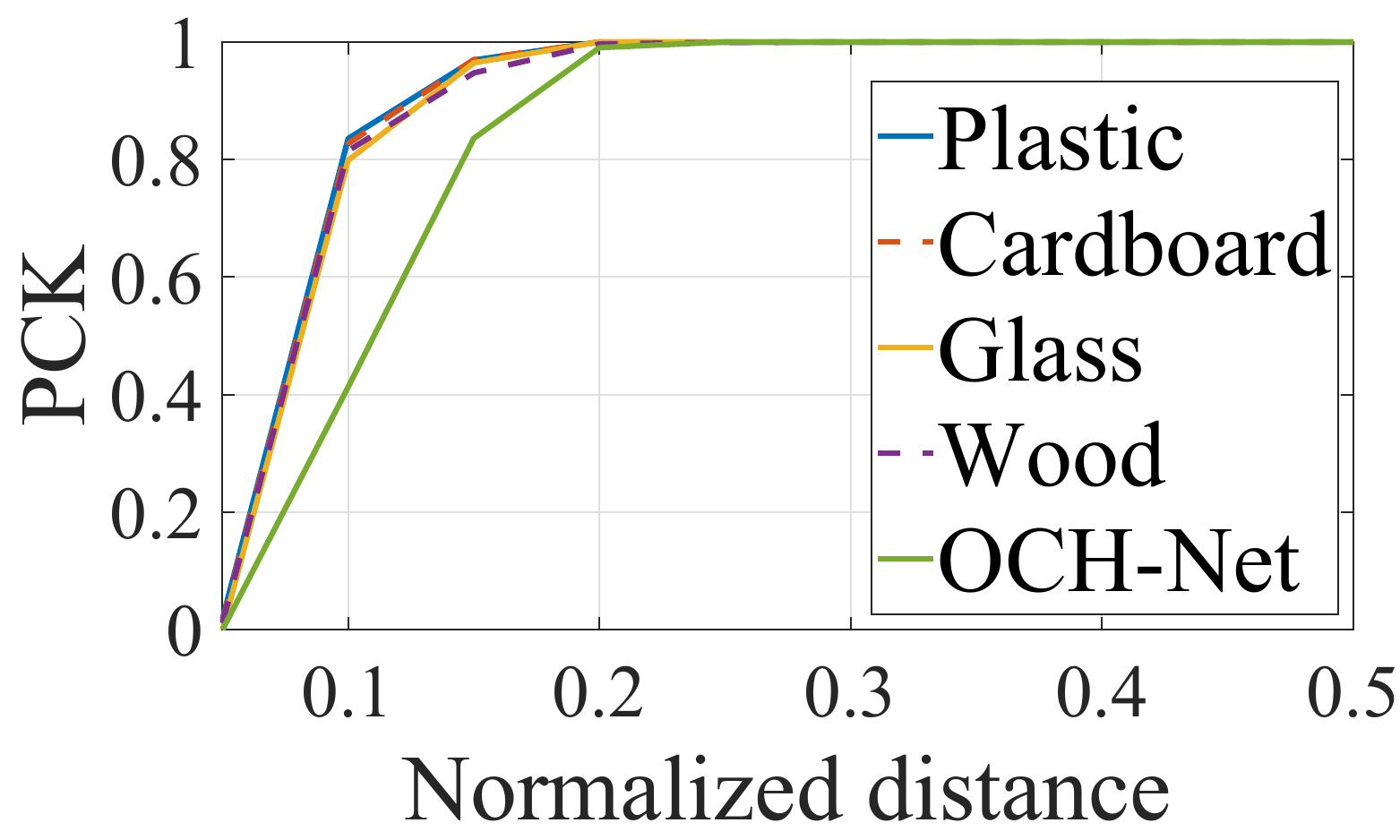}
                \vspace{-2.5ex}
			\label{sfig:unseen_mcp}
		\end{minipage}
	}
	\subfloat[{PIP.}]{ 
		\begin{minipage}[b]{.48\linewidth}
			\centering
			\includegraphics[width = \textwidth]{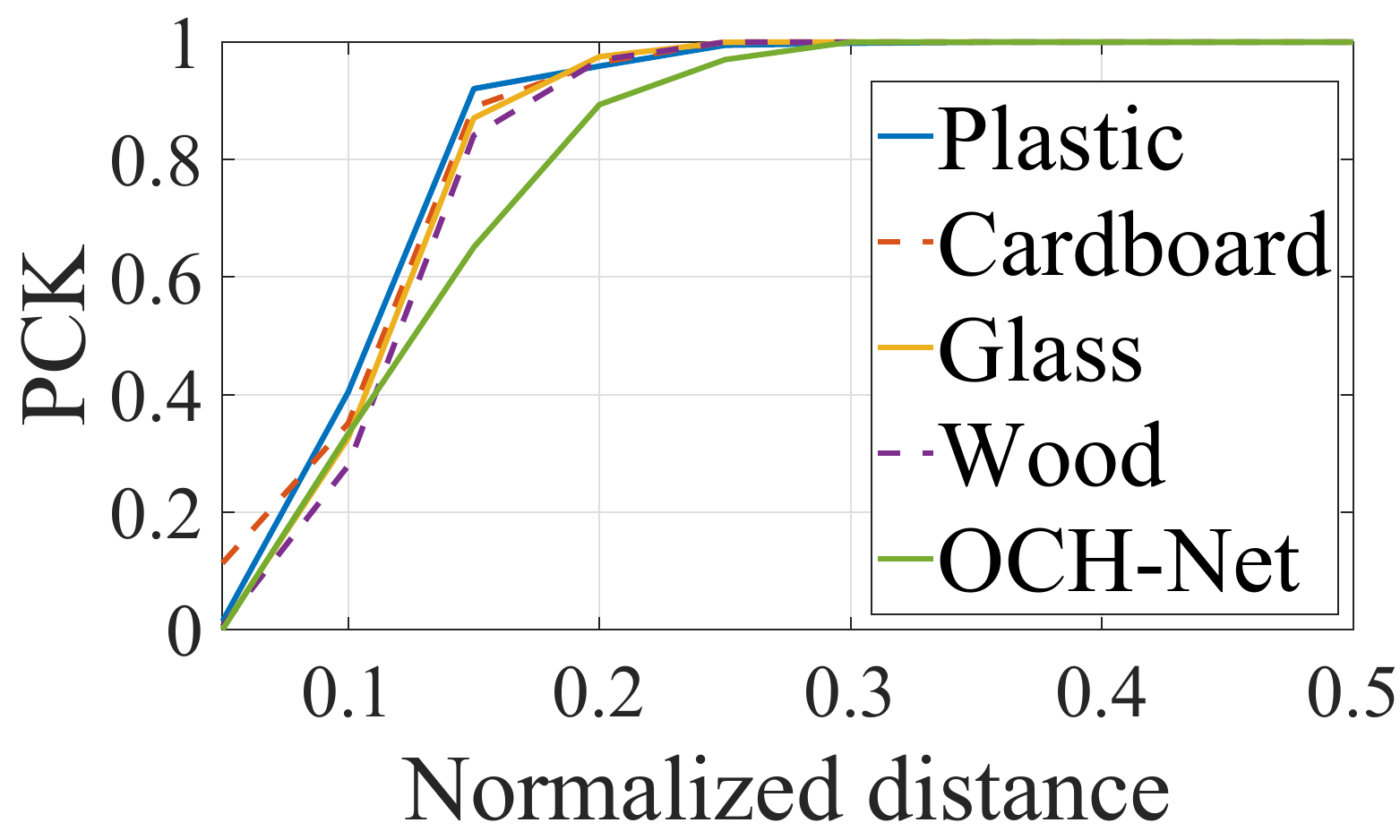}
                \vspace{-2.5ex}
			\label{sfig:unseen_pip}
		\end{minipage}
	}
        \\ \vspace{-2ex}
	\subfloat[DIP.]{ 
		\begin{minipage}[b]{.48\linewidth}
			\centering
			\includegraphics[width = \textwidth]{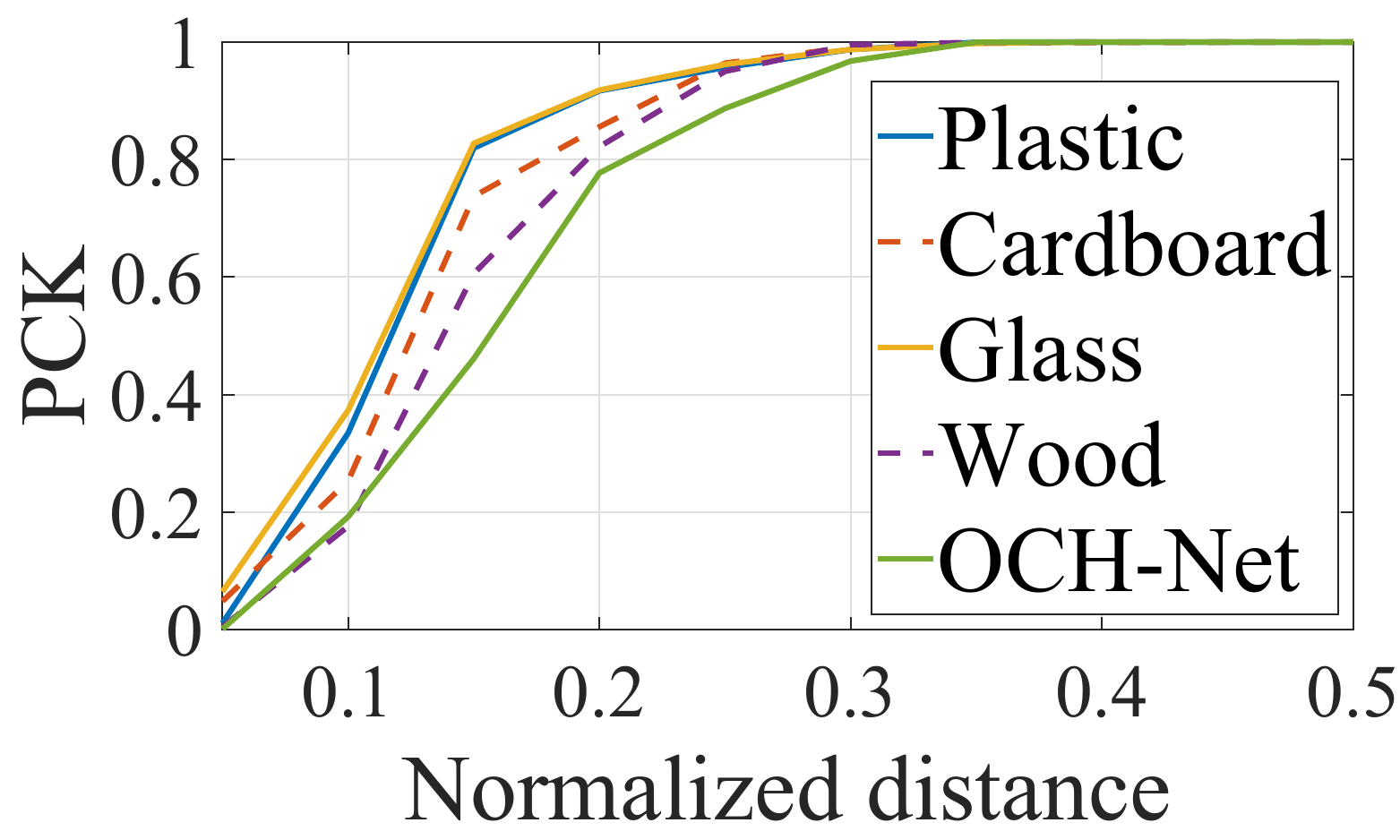}
                \vspace{-2.5ex}
			\label{sfig:unseen_dip}
		\end{minipage}
	}
	\subfloat[Fingertip.]{ 
		\begin{minipage}[b]{.48\linewidth}
			\centering
			\includegraphics[width = \textwidth]{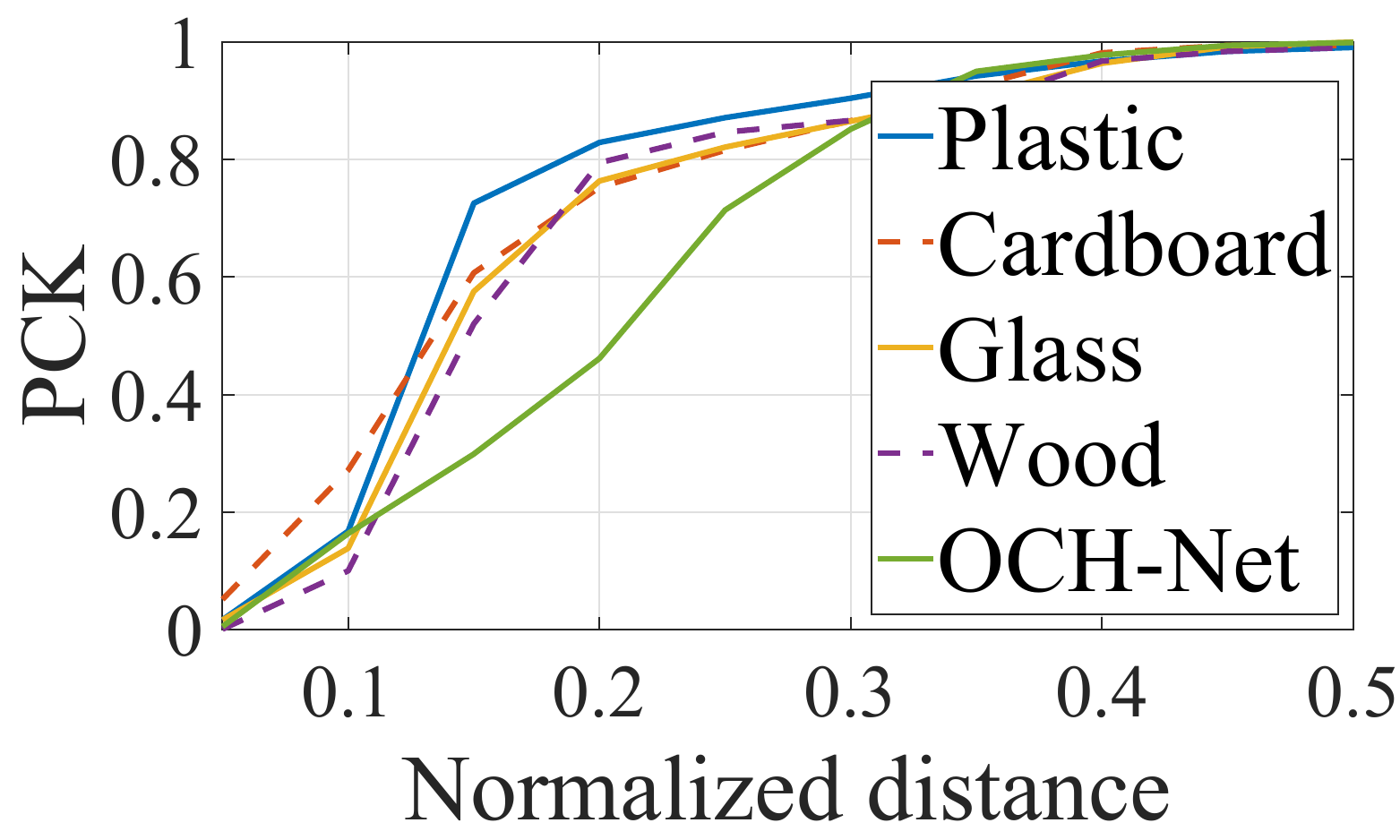}
                \vspace{-2.5ex}
			\label{sfig:unseen_fingertip}
		\end{minipage}
	}
        \vspace{1ex}
	\caption{Comparing PCKs of \nnname and the average PCKs of \nname under unseen occlusion.}
	\label{fig:finger_ad_unseen}
        \vspace{-2.5ex}
\end{figure}

To validate the generalizability of \nnname, we conduct additional experiments to assess its performance on unseen obstacles during adaptation. Specifically, we select three types of obstacles from wood, plastic, glass, and cardboard sheets for adaptation, and leave one out for testing. The results presented in Figure~\ref{fig:finger_ad_unseen} show the PCKs@0.2 values achieved by \nnname for various keypoints. Our findings suggest that \nnname yields remarkable PCKs@0.2 scores far exceeding those of the average PCKs of \nname. Furthermore, our experiments reveal that \nnname performs slightly better when the obstacle is plastic or cardboard than when it is glass or wood, possibly due to the lower dielectric and loss tangent of these materials. These results strongly demonstrate the generalizability of \nnname to unseen domains, which is a significant advantage over methods that require time-consuming and cumbersome retraining.

\begin{figure}[b]
    \vspace{-1ex}
    \centering
    \begin{minipage}[b]{.49\linewidth}
    \includegraphics[width = \textwidth]{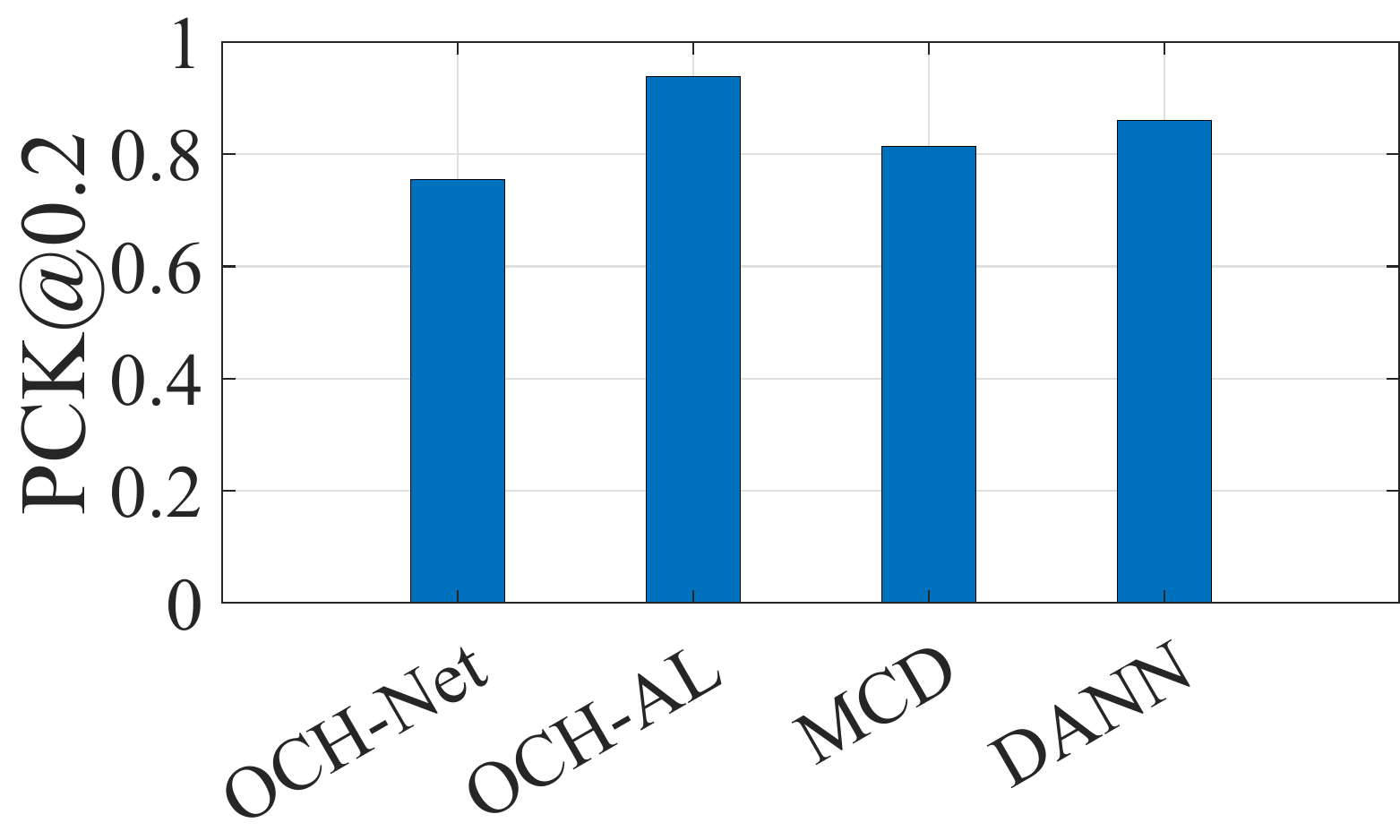}
    \caption{Different adversarial learning algorithms.}
    \label{fig:diff_al}
    \end{minipage}
    \hfill
    \begin{minipage}[b]{.49\linewidth}
    \includegraphics[width = \textwidth]{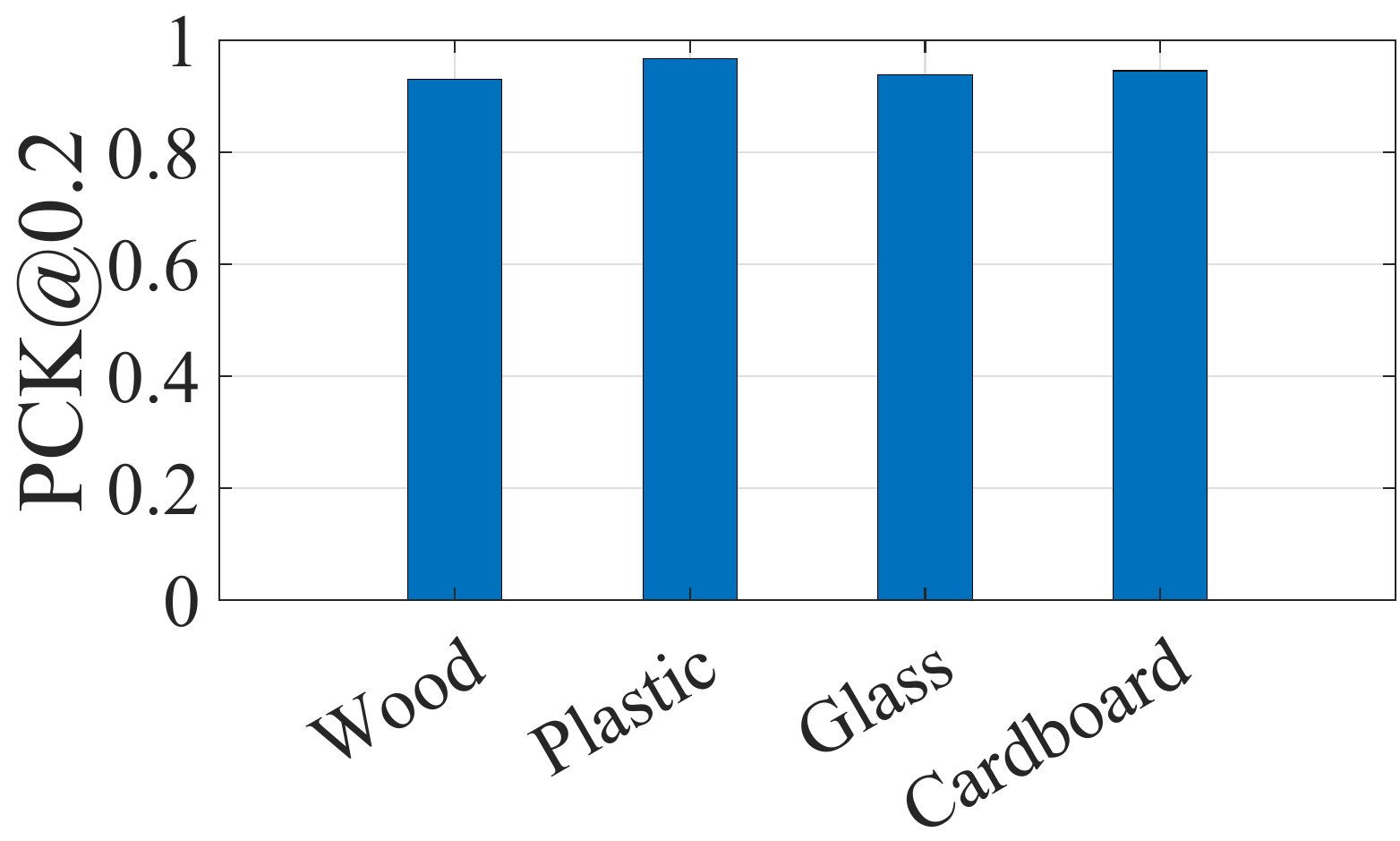}
    \caption{Different obstacle materials.}
    \label{fig:block_materials}
    \end{minipage}
    
\end{figure}

\vspace{-2ex}
\subsubsection{Adversarial Learning Algorithm Comparison}
We compare our \nnname with two other adversarial learning algorithms DANN~\cite{long2018conditional} and MCD~\cite{saito2018maximum}. To ensure a fair comparison, we use the same feature extractor in all three algorithms. We calculate the average PCKs@0.2 for each algorithm and show them in Figure~\ref{fig:diff_al}. All three algorithms improve the original \nname’s performance, but our \nnname achieves the highest PCK@0.2. Moreover, we observe that while MCD and DANN help \nname adapt to occlusion, their average PCKs@0.2 in the normal domain decrease by 0.10 and 0.13, respectively. In contrast, \nnname maintains consistent performance in both normal and occluded domains. As described in Section~\ref{ssec:dal}, \nnname is specifically designed for the HPE regression task by bounding the size of the output space, so it achieves successful domain adaptation while avoiding unnecessary parameter updates, thus preserving \nname's performance in the normal domain. 

\vspace{-2.5ex}
\subsubsection{Ablation Study}
We hereby evaluate the different components of \nname, and study their impact on the performance.  There are three baselines called \textit{Real-Net}, \textit{I-Net}, and \textit{Q-Net}. The Real-Net is implemented by removing all the complex-valued CNN operations designed in Section~\ref{ssec:dcvn}, and simply concatenating the I and Q branches. The I-Net and Q-Net use the same structure as Real-Net, but are trained only on the I or Q RF data, respectively. In this experiment, we consider only data from the normal domain.
\vspace{-1ex}
\begin{figure}[h]
    \vspace{-1.5ex}
	\centering
	\subfloat[MCP.]{ 
		\begin{minipage}[b]{.48\linewidth}
			\centering
			\includegraphics[width = \textwidth]{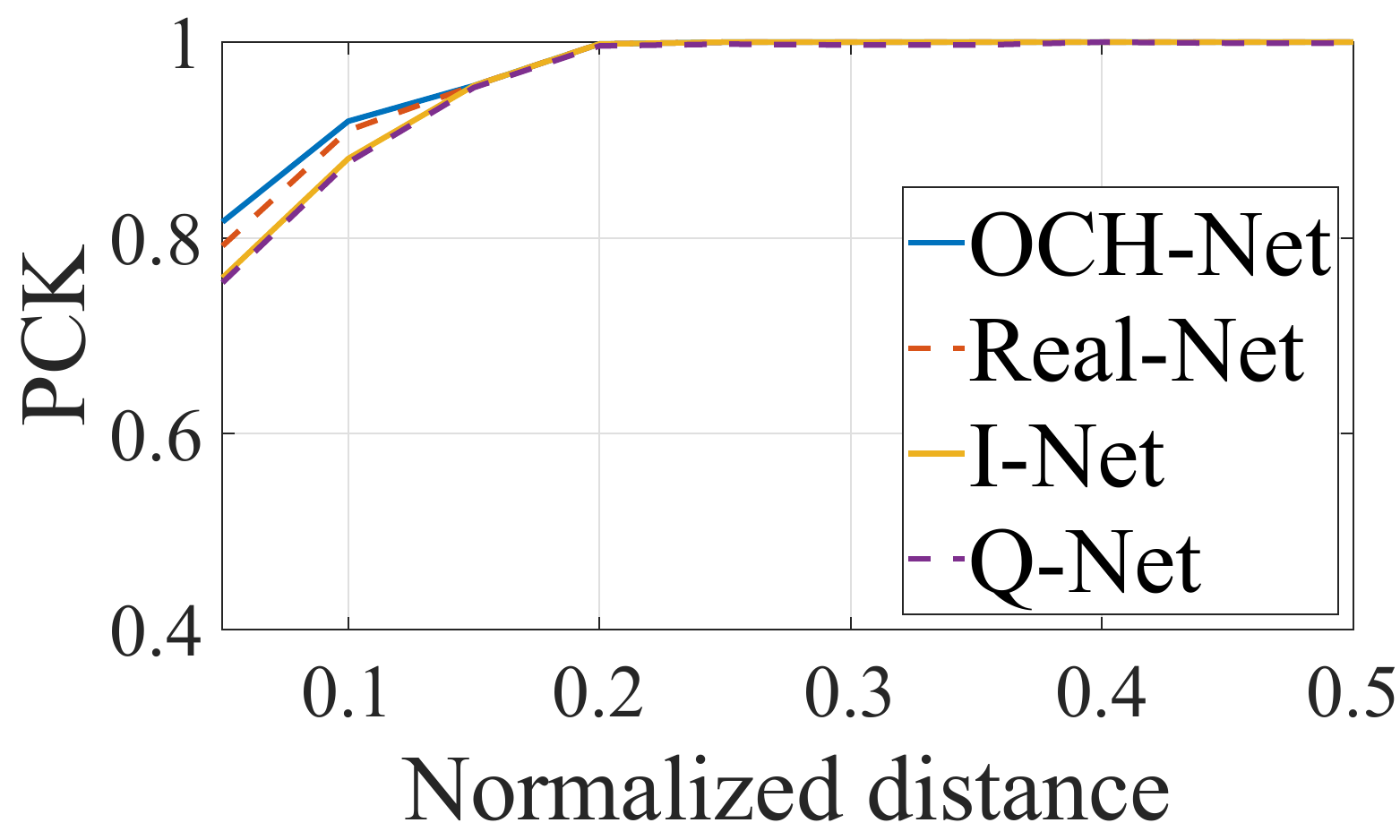}
			\label{sfig:mb_mcp}
            \vspace{-2ex}
		\end{minipage}
	}
	\subfloat[{PIP.}]{ 
		\begin{minipage}[b]{.48\linewidth}
			\centering
			\includegraphics[width = \textwidth]{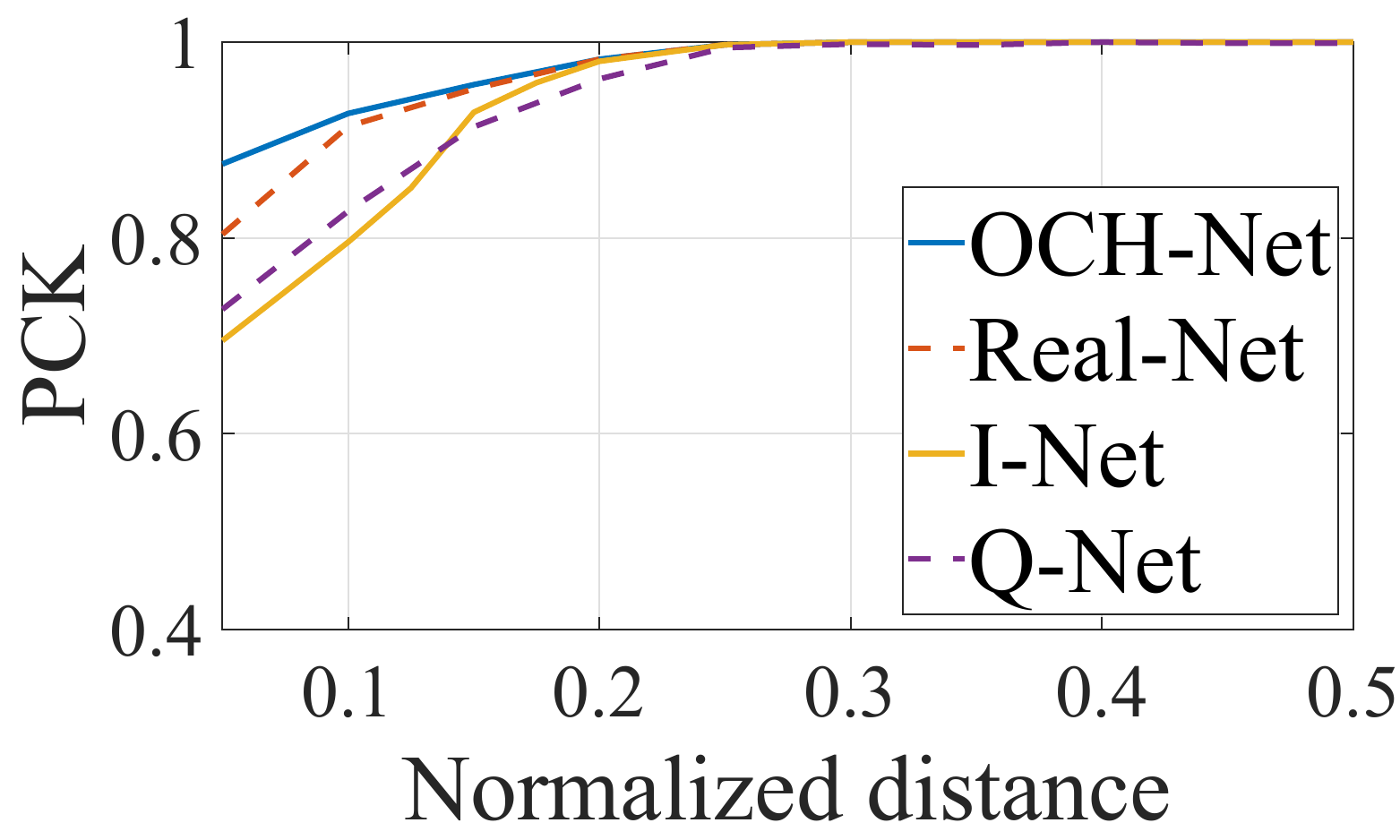}
			\label{sfig:mb_pip}
            \vspace{-2ex}
		\end{minipage}
	} \\\vspace{-2ex}
		\subfloat[DIP.]{ 
		\begin{minipage}[b]{.48\linewidth}
			\centering
			\includegraphics[width = \textwidth]{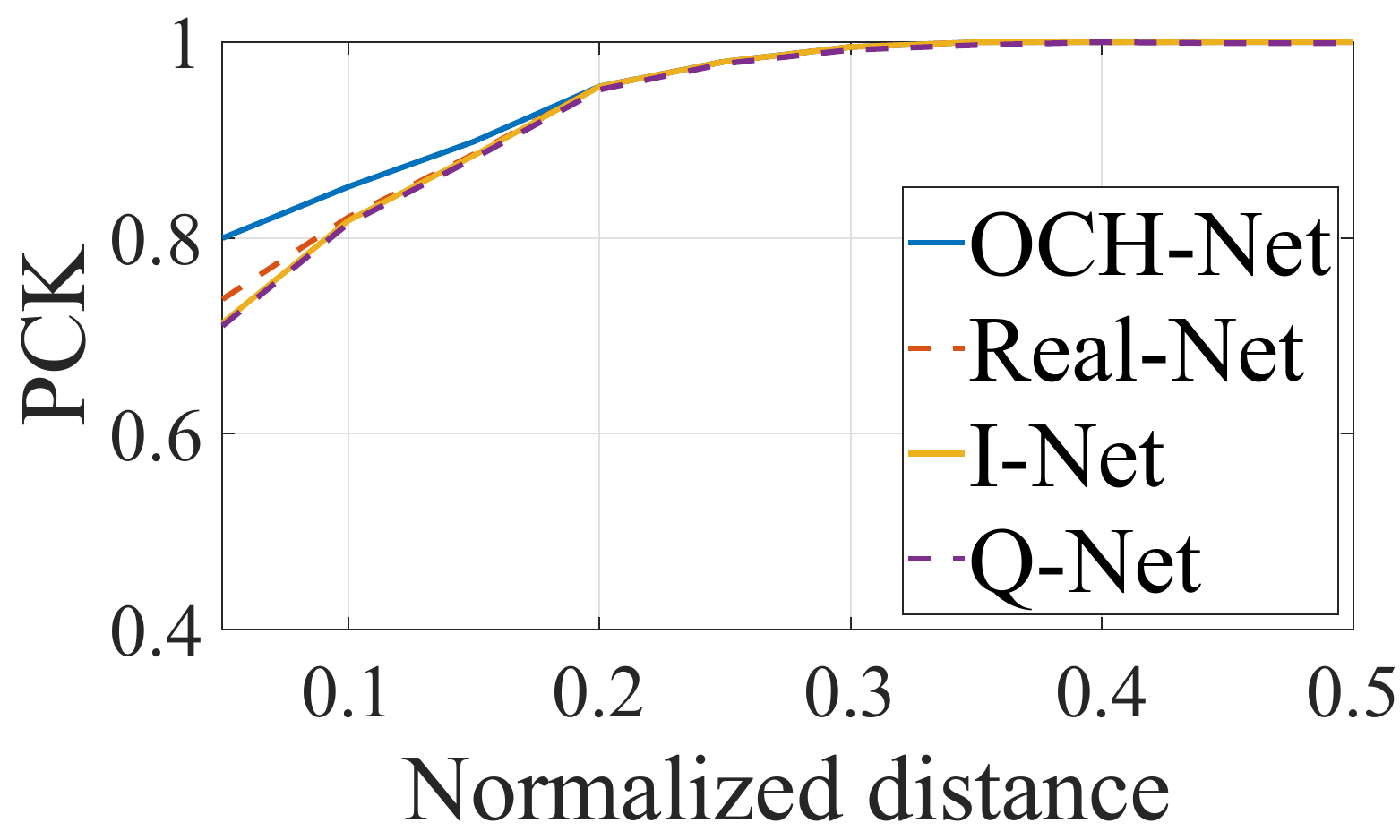}
			\label{sfig:mb_dip}
            \vspace{-2ex}
		\end{minipage}
	}
		\subfloat[Fingertip.]{ 
		\begin{minipage}[b]{.48\linewidth}
			\centering
			\includegraphics[width = \textwidth]{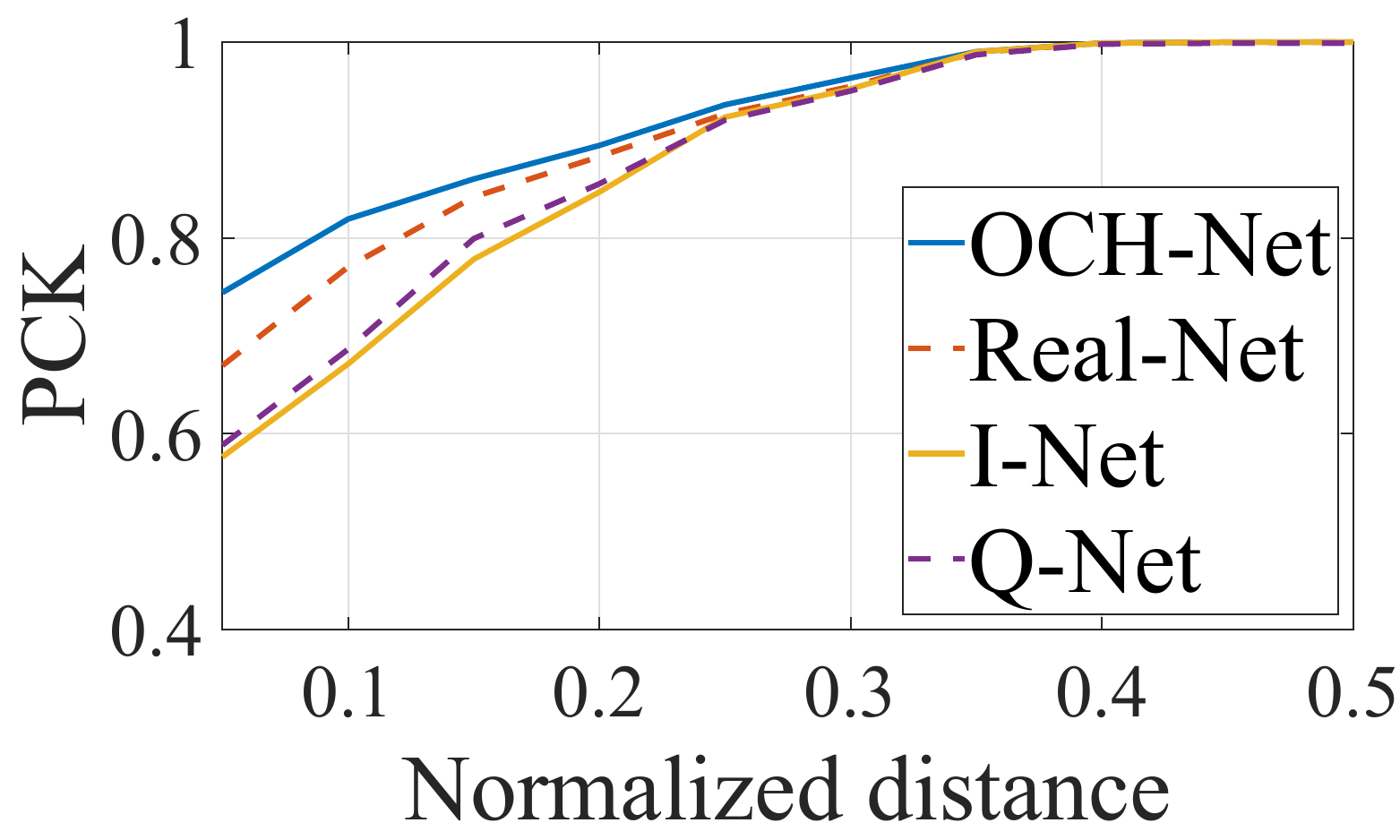}
			\label{sfig:mb_fingertip}
            \vspace{-2ex}
		\end{minipage}
	}
     \vspace{1ex}
	\caption{PCKs of different neural network structures.}
	\label{fig:finger_mb}
     \vspace{-2ex}
\end{figure}

Apparently, \nname outperforms all three baselines for all parts of the hand.  The PCKs@0.2 of \nname are 0.9988,  0.9823, 0.9546, and  0.8943 for MCP, PIP, DIP, and fingertip, respectively. 
Due to the power of complex-valued CNN operations, \nname obtains noticeable improvement over Real-Net, I-Net, and Q-Net. Moreover, I-Net and Q-Net achieve similar PCKs@0.2 for all parts of the hand, and are consistently the worst among all baselines. The reason is that single I or Q neural network structure cannot represent the whole RF data intrinsically. Moreover, it can be observed that PCKs decrease slightly from Figure~\ref{sfig:mb_mcp} to Figure~\ref{sfig:mb_fingertip}, probably because the motion-induced errors from MCP to fingertip grows larger progressively. 

\vspace{-2ex}
\subsubsection{Impact of Different Obstacle Materials } %
We investigate the impact of obstacle materials including wood, plastic, glass, and cardboard on the performance of our system. To achieve this, we place these materials in front of the RF sensor to block all LoS RF signals. As shown in Figure~\ref{fig:block_materials}, the average PCKs@0.2 are found to be 0.9309, 0.9672, 0.9381, and 0.9464 for wood, plastic, glass, and cardboard, respectively. Notably, the worst performance is observed with the wood block, which causes the largest \newrev{interference to RF signals among the four obstacles, thereby altering the RF signal distribution reflected by a hand to the most extent.}
\begin{figure}[t]
    \centering
    \begin{minipage}[b]{.49\linewidth}
    \includegraphics[width = \textwidth]{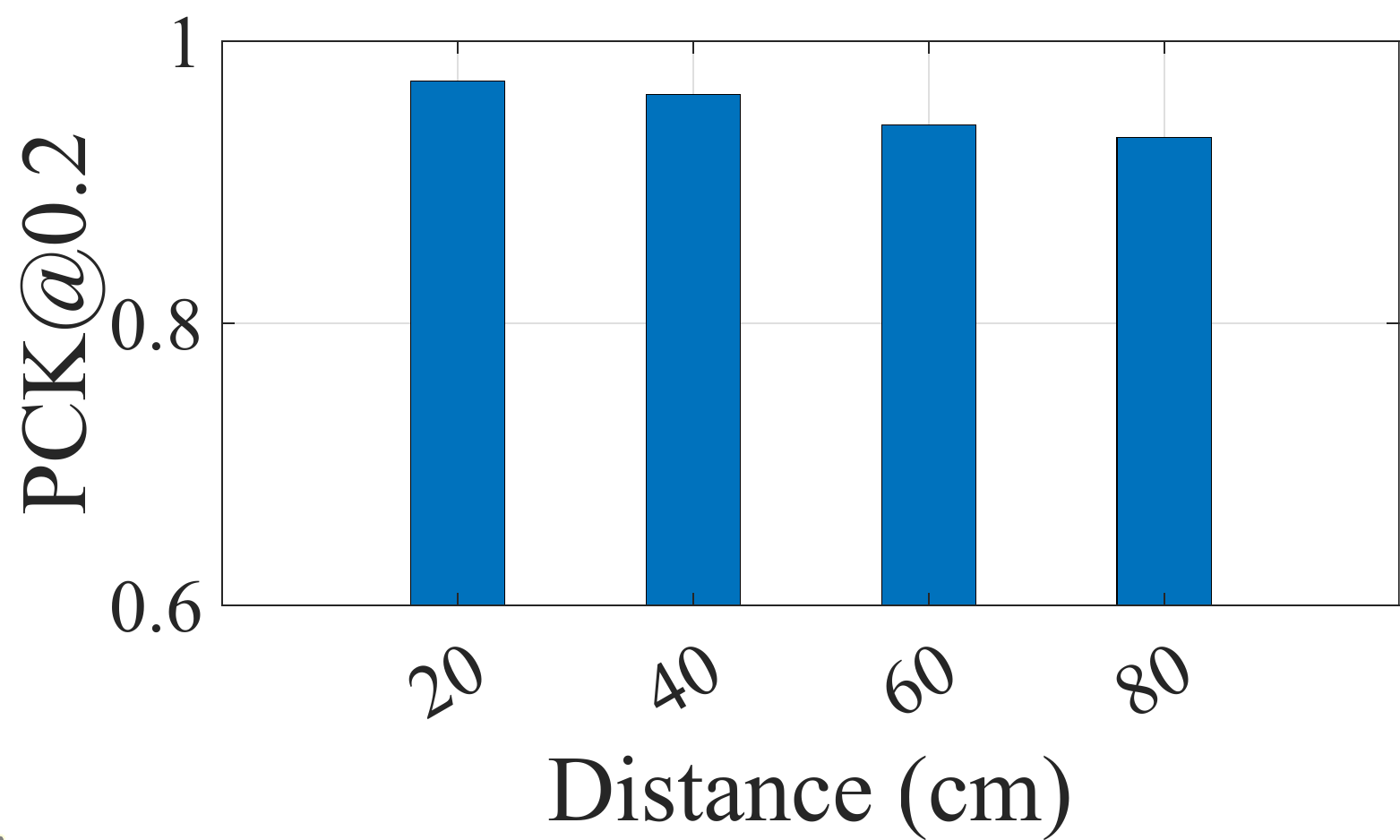}
    \caption{Different distances.}
    \label{fig:dis_pck}
    \end{minipage}
    \hfill
    \begin{minipage}[b]{.49\linewidth}
    \includegraphics[width = \textwidth]{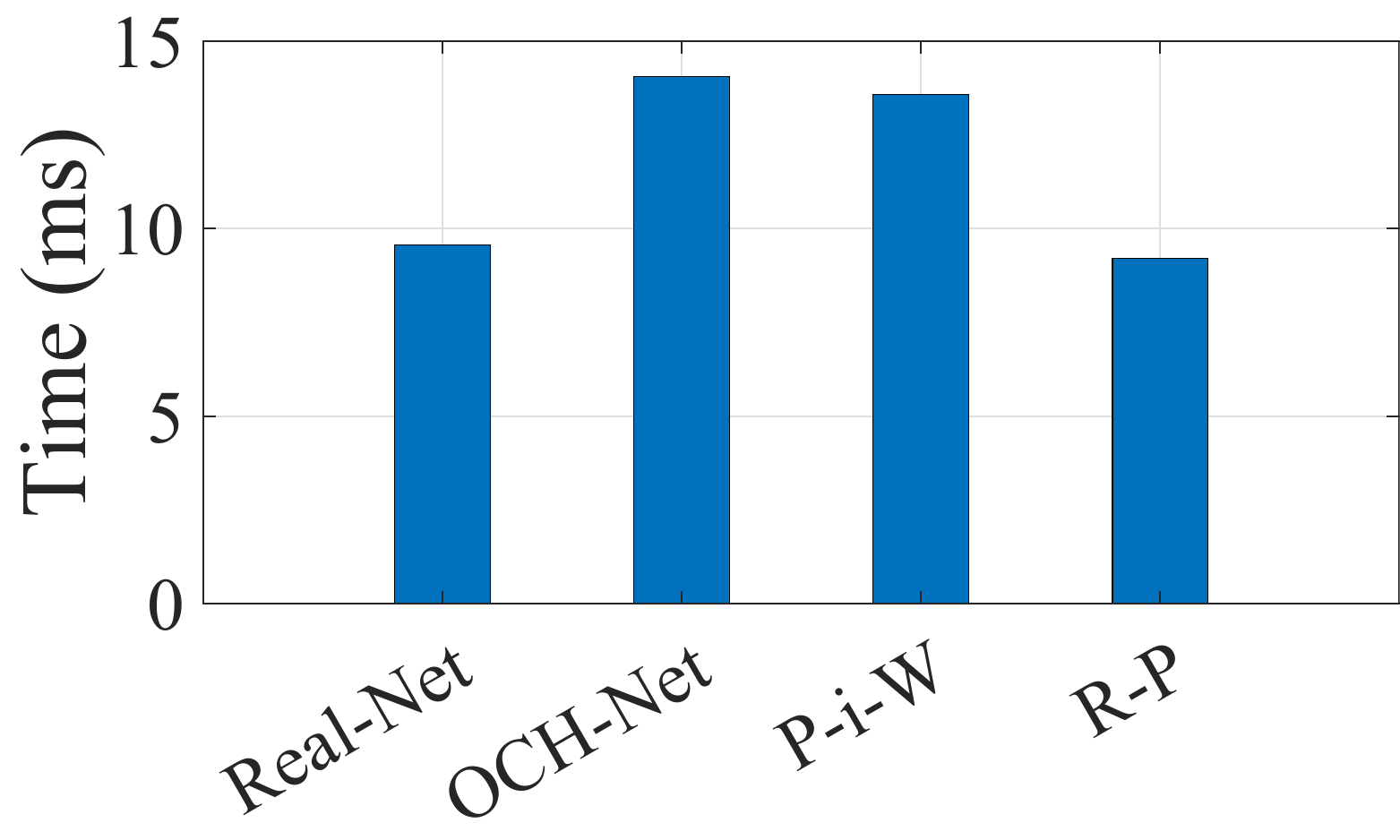}
    \caption{Inference time.}
    \label{fig:speed_nn}
    \end{minipage}
\vspace{-2.5ex}
\end{figure}

\vspace{-2ex}
\subsubsection{Impact of Different Distances}
To evaluate the impact of different distances, we test the trained \name with data containing hands at different distances from the RF sensor. The results, depicted in Figure~\ref{fig:dis_pck}, reveal that the average PCKs@0.2 decreases by 4\% as the distance increases from 20~\!cm to 80~\!cm. This degradation can be explained by the fact that the RF sensor captures more interference as the distance increases. Moreover, the receiving power of RF signals decreases with distance, resulting in lower SNRs in reflected RF signals. Despite the slight performance degradation, \name can still provide sufficient HPE accuracy at a further distance. 

\vspace{-2ex}
\subsubsection{Inference Time of \name}
We further evaluate the inference time of \name. In the inference stage, we only keep the feature extractor $\phi$ and the normal regressor $g$ for assessment. We compare the inference time of \nname with three baselines, and show the average inference time in Figure~\ref{fig:speed_nn}. The average inference time of \nname is 14~\!ms, which is only slightly higher than those of the baselines by up to 5~\!ms. The main reason for the extra computing overhead is the use of complex-valued neural networks. If we replace the \nname with Real-Net, the inference time is decreased to 10~\!ms. However, we believe that the modest overhead of only 4~\!ms is worth the superior performance that \nname provides for our HPE task.

\section{Conclusion} \label{sec:con} %
HPE in occluded scenarios is a crucial yet challenging problem pertinent to human-computer interaction. 
In this paper, to overcome the LoS limitations of CM-HPE, we resort to RF-vision, and propose \name as a cross-modality, cross-domain method for occlusion-robust HPE. Employing the carefully designed cross-modality framework, we have demonstrated \name's ability to map RF signals to hand keypoints in a non-Euclidean manner.  Furthermore, \name has successfully adapted its neural model \nname to deal with diversified obstacles by leveraging the power of adversarial learning. 
\camrev{Extensive experiments have been conducted to demonstrate that \name achieves high accuracy in HPE, even in occluded scenarios. The results strongly support the effectiveness of this method, showing its potential for practical applications in fields such as human-computer interaction (HCI), smart-home controls, and medical rehabilitation.}

{\small
\bibliographystyle{ieee_fullname}
\bibliography{ref}
}
\clearpage
\appendix
\end{document}